\documentclass[lettersize,journal]{IEEEtran}
\usepackage{amsmath,amsfonts}
\usepackage{algorithmic}
\usepackage{algorithm}
\usepackage{array}
\usepackage[caption=false,font=normalsize,labelfont=sf,textfont=sf]{subfig}
\usepackage{textcomp}
\usepackage{stfloats}
\usepackage{url}
\usepackage{verbatim}
\usepackage{graphicx}
\usepackage{cite}
\usepackage{xspace}
\usepackage{colortbl}
\usepackage{booktabs}
\usepackage{float}
\usepackage{enumitem}
\usepackage{multirow}
\usepackage{wrapfig, capt-of}
\usepackage[misc]{ifsym} 

\def\BibTeX{{\rm B\kern-.05em{\sc i\kern-.025em b}\kern-.08em
    T\kern-.1667em\lower.7ex\hbox{E}\kern-.125emX}}

\usepackage{amsmath,amsfonts,bm}

\def\eqref#1{equation~\ref{#1}}

\def\1{\bm{1}}

\def\rvc{{\mathbf{c}}}
\def\rvd{{\mathbf{d}}}

\def\rvp{{\mathbf{p}}}

\def\rvr{{\mathbf{r}}}

\DeclareMathAlphabet{\mathsfit}{\encodingdefault}{\sfdefault}{m}{sl}
\SetMathAlphabet{\mathsfit}{bold}{\encodingdefault}{\sfdefault}{bx}{n}

\def\gN{{\mathcal{N}}}

\newcommand{\set}[1]{\mathcal{#1}}

\newcommand{\loss}{\mathcal{L}}

\def\onedot{\xspace}

\def\eg{\emph{e.g}\onedot} 
\def\ie{\emph{i.e}\onedot}

\def\etal{\emph{et al}\onedot}

\usepackage[dvipsnames]{xcolor}
\definecolor{magenta(process)}{rgb}{1.0, 0.0, 0.9}
\newcommand{\heading}[1]{\noindent\textbf{#1.}}

\newcommand{\NICKNAME}{\textsc{LN3Diff++}} 
\newcommand{\nickname}{\NICKNAME} 

\newcommand{\image}{I} 

\newcommand{\real}{\mathbb{R}}

\newcommand{\RN}[1]{%
  \textup{\uppercase\expandafter{\romannumeral#1}}%
}

\newcommand{\diffrender}{\mathbf{R}}

\definecolor{yellow}{rgb}{1, 1, 0.7}
\definecolor{orange}{rgb}{1, 0.85, 0.7}
\definecolor{tablered}{rgb}{1, 0.7, 0.7}
\definecolor{red}{rgb}{1, 0, 0}

\newcommand{\cm}[1]{{\color{black}{#1}}} 
\newcommand{\cmfinal}[1]{{\color{black}{#1}}} 

\newcommand{\expec}{\mathbb{E}}

\newcommand{\lsimplediff}{\loss_\text{diff}}

\newcommand{\model}{\beps_\theta}
\newcommand{\modelwithinp}{\model(\bz_t, t)}

\newcommand{\ldmencoder}{\mathcal{E}_\bphi}
\newcommand{\ldmdecoder}{\mathcal{D}_\bpsi}

\newcommand{\bz}{{\mathbf{z}}}
\newcommand{\by}{{\mathbf{y}}}

\newcommand{\beps}{\boldsymbol{\epsilon}}

\newcommand{\btheta}{{\boldsymbol{\theta}}}
\newcommand{\bphi}{{\boldsymbol{\phi}}}
\newcommand{\bpsi}{{\boldsymbol{\psi}}}
\newcommand{\balpha}{{\boldsymbol{\alpha}}}
\newcommand{\bzero}{\mathbf{0}}
\newcommand{\beye}{\mathbf{I}}

\newcommand{\N}{\mathcal{N}}
\newcommand{\U}{\mathcal{U}}

\usepackage[capitalize]{cleveref}
\crefname{section}{Sec.}{Secs.}
\Crefname{section}{Section}{Sections}
\Crefname{table}{Table}{Tables} 
\crefname{table}{Tab.}{Tabs.}

\begin{document}

\title{\nickname{}: Scalable \textbf{L}atent \textbf{N}eural Fields \textbf{D}iffusion for Speedy \textbf{3D} Generation}


\author{Yushi Lan,~\IEEEmembership{Student Member,~IEEE},  
Fangzhou Hong,
Shangchen Zhou,
Shuai Yang,~\IEEEmembership{Member,~IEEE},
Xuyi Meng,
Yongwei Chen,
Zhaoyang Lyu,
Bo Dai,
Xingang Pan,
Chen Change Loy $^\text{\Letter}$,~\IEEEmembership{Senior Member,~IEEE}, 
}

\markboth{Journal of \LaTeX\ Class Files,~Vol.~18, No.~9, September~2020}%
{How to Use the IEEEtran \LaTeX \ Templates}



\maketitle

\begin{abstract}
The field of neural rendering has seen remarkable progress, driven by advancements in generative models and differentiable rendering techniques. While 2D diffusion has achieved notable success, the development of a unified 3D diffusion pipeline remains an open challenge. 
This paper presents a novel framework, \textbf{\nickname{}}, designed to bridge this gap and facilitate fast, high-quality, and versatile conditional 3D generation. Our method leverages a 3D-aware architecture and a variational autoencoder (VAE) to encode input image(s) into a structured, compact 3D latent space. The latent representation is then decoded by a transformer-based decoder into a high-capacity 3D neural field.
By training a diffusion model on this 3D-aware latent space, our method achieves superior performance for category-specific 3D generation on ShapeNet and FFHQ, as well as category-free image/text-conditioned 3D generation over Objaverse. Moreover, it surpasses existing 3D diffusion methods in inference speed, requiring no per-instance optimization. 
Video demos can be found on our project webpage: 
\url{https://nirvanalan.github.io/projects/ln3diff}.
\end{abstract}
\begin{figure*}[h!]
\centering{
\includegraphics[width=1.0\textwidth]{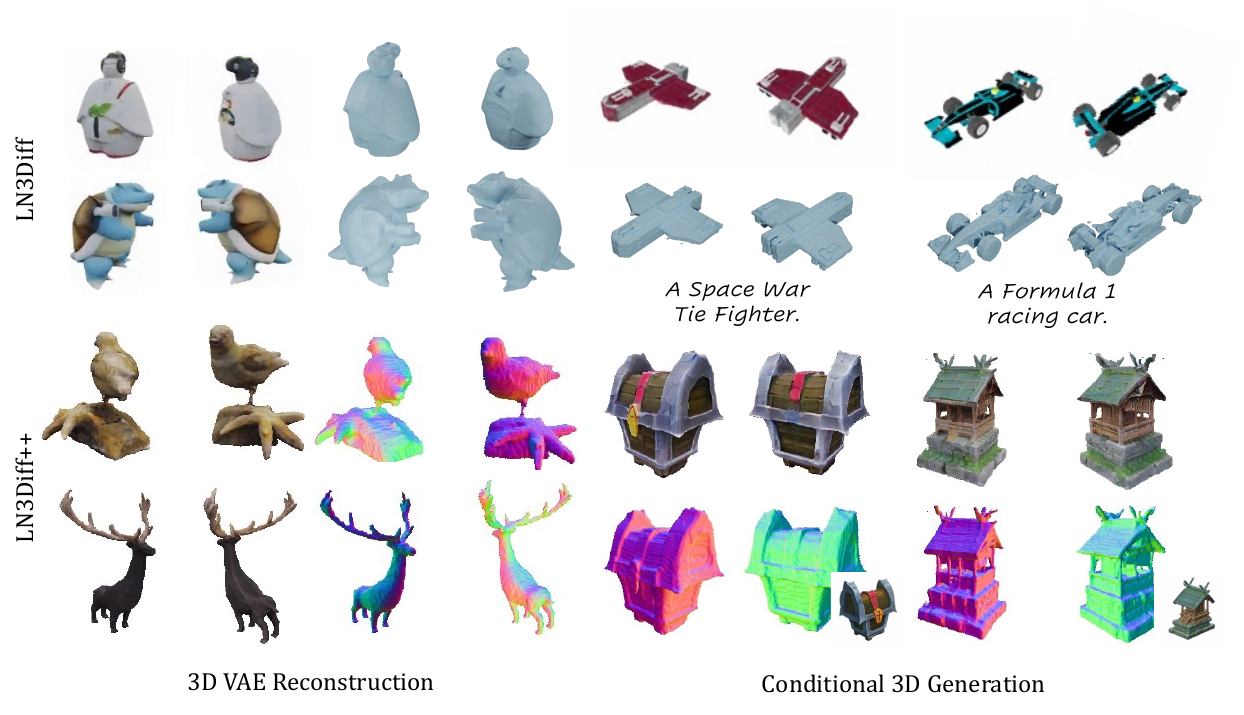}
\caption{We present \nickname{}, which performs efficient 3D diffusion learning over a compact latent space.
\cm{Compared to LN3Diff which adopts NeRF rendering and supports text-conditioned 3D generation, \nickname{} further enables SDF-based 3D representation and image-conditioned 3D generation.}
The resulting model enables both high-quality monocular 3D reconstruction and text-to-3D synthesis.
}
\label{fig:teaser}
}
\end{figure*}

\section{Introduction}
\label{sec:intro}

\IEEEPARstart{T}{he} advancement of generative models~\cite{Jo2022DDPM,Goodfellow2014GenerativeAN} and differentiable rendering~\cite{Tewari2021AdvancesIN} has paved the way for a new research direction called neural rendering~\cite{Tewari2021AdvancesIN}. This field is continuously pushing the limits of view synthesis~\cite{mildenhall2020nerf}, editing~\cite{lan2022ddf_ijcv}, and particularly 3D object synthesis~\cite{Chan2021EG3D}. While 2D diffusion models~\cite{song2021scorebased,Jo2022DDPM} have outperformed generative adversarial networks (GANs)-based methods in image synthesis~\cite{dhariwal2021diffusion} in quality~\cite{rombach2022LDM}, controllability~\cite{zhang2023adding}, and scalability~\cite{Schuhmann2022LAION5BAO}, a unified 3D diffusion pipeline has yet to be established.

3D object generation methods using diffusion models can be categorized into 2D-lifting and feed-forward 3D diffusion models.
In 2D-lifting methods, score distillation sampling (SDS)~\cite{poole2022dreamfusion,wang2023prolificdreamer} and Zero-123~\cite{liu2023zero1to3,shi2023zero123plus} achieve 3D generation by leveraging pre-trained 2D diffusion models. However, SDS-based methods require costly per-instance optimization and are prone to the multi-face Janus problem~\cite{poole2022dreamfusion}, where models struggle to maintain consistency and fidelity across different views of the same face. Meanwhile, Zero-123 fails to enforce strict view consistency.
On the other hand, feed-forward 3D diffusion models~\cite{wang2023rodin,muller2023diffrf,Jun2023ShapEGC,lan2023gaussian3diff,ssdnerf,3dshape2vecset} enable fast 3D synthesis without per-instance optimization. However, these methods typically involve a two-stage pre-processing approach. First, during the data preparation stage, a shared decoder is learned over a large number of instances to ensure a shared latent space. This is followed by per-instance optimization to convert each 3D asset in the datasets into neural fields~\cite{Xie2021NeuralFI}. After this, the feed-forward diffusion model is trained on the prepared neural fields.

While the pipeline above is straightforward, it poses extra challenges to achieve high-quality 3D diffusion:
\textbf{1)~Scalability.}
In the data preparation stage,
existing methods face scalability issues due to the use of a shared, low-capacity MLP decoder for per-instance optimization.
This approach is data inefficient, requiring over $50$ views per instance~\cite{muller2023diffrf,ssdnerf} during training.
Consequently, the computation cost scales linearly with the size of the dataset, hindering scalability for large, diverse 3D datasets.
\textbf{2)~Efficiency.}
Employing 3D-specific architectures~\cite{qi2016pointnet,thomas2019kpconv,spconv2022} is computationally intensive and necessitates representation-specific designs~\cite{Zhou2021PVD}. Consequently, existing methods compress each 3D asset into neural fields~\cite{Xie2021NeuralFI} before training. However, this compression introduces high-dimensional 3D latent, increasing computational demands and training challenges. Limiting the neural field size~\cite{ssdnerf} might mitigate these issues but at the cost of reconstruction quality. In addition, the auto-decoding paradigm can result in an unclean latent space~\cite{Shue20223DNF,gu2023control3diff,Jun2023ShapEGC}, unsuitable for 3D diffusion training~\cite{rombach2022LDM}.
\textbf{3)~Generalizability.} 
Existing 3D diffusion models primarily focus on unconditional generation over single classes, neglecting high-quality conditional 3D generation (\eg, text-to-3D) across generic category-free 3D datasets.
Furthermore, projecting monocular input images into the diffusion latent space is crucial for conditional generation and editing~\cite{zhang2023adding,lan2022e3dge}, but this is challenging with the shared decoder designed for multi-view inputs.


In this study, we propose a novel framework called {\textbf{L}atent \textbf{N}eural fields \textbf{3}D \textbf{Diff}usion (\nickname{})} to address these challenges and enable fast, high-quality and generic conditional 3D generation.
%
Our method involves training a variational autoencoder~\cite{Kingma2013AutoEncodingVB} (VAE) to compress input images into a lower-dimensional 3D-aware latent space, which is more expressive and flexible compared to pixel-space diffusion~\cite{song2021scorebased,Hoogeboom2023simpleDE,Jo2022DDPM,dhariwal2021diffusion}.
From this space,
a 3D-aware transformer-based decoder gradually decodes the latent into a high-capacity 3D neural field.
This autoencoding stage is trained amortized with differentiable rendering~\cite{Tewari2021AdvancesIN}, incorporating novel view supervision for multi-view datasets~\cite{shapenet2015,objaverse} and adversarial supervision for monocular datasets~\cite{karras2018progressive}.
Thanks to the high-capacity model design, our method is more \emph{view efficient}, requiring only two views per instance during training.
After training, we leverage the learned 3D latent space for conditional 3D diffusion learning, ensuring effective utilization of the trained model for high-quality 3D generation.
The pre-trained encoder can amortize the data encoding over incoming data, thus streamlining operations and facilitating efficient 3D diffusion learning while remaining compatible with advances in 3D representations.

To enhance efficient information flow in the 3D space and promote coherent geometry reconstruction, we introduce a novel 3D-aware architecture tailored for fast and high-quality 3D reconstruction while maintaining a structured latent space.
Specifically, we employ a convolutional tokenizer to encode the input image(s) into a $\emph{KL}$-regularized 3D latent space, leveraging its superior perceptual compression ability~\cite{esser2020taming}.
We employ transformers~\cite{Peebles2022DiT,dosovitskiy2020vit} to enable flexible 3D-aware attention across 3D tokens in the latent space.
Finally, we up-sample the 3D latent and apply differentiable rendering for image-space supervision, making our method a self-supervised 3D learner~\cite{sitzmann2019srns}. 
%

An earlier version of this work appeared in Lan~\etal~\cite{lan2024ln3diff}, which focuses on unconditional generation over ShapeNet and preliminary study over category-free text-conditioned 3D generation. Besides, U-Net and DDPM framework is adopted following the conventions~\cite{dhariwal2021diffusion}.
\cmfinal{
To further improve the flexibility of 3D generation, this journal extension further explores the architecture of image-conditioned 3D generation based on the modern diffusion transformer (DiT)~\cite{Peebles2022DiT} architecture. Besides, the well-established flow matching (FM)~\cite{albergo2023stochastic,lipman2022,liu2022flow} framework is incorporated for scalable diffusion training. 
Motivated by recent evidence that pairing FM with Diffusion Transformers (DiT) advances video generation~\cite{wan2025}, we replace the DDPM+U-Net paradigm with FM+DiT. This change improves both generation quality and training efficiency in our method and situates our 3D generative model within a unified image–video framework, potentially enabling joint reasoning and generation as demonstrated in state-of-the-art image diffusion models~\cite{chen2025blip3ofamilyfullyopen,pan2025transfer}.
Additionally, to enable high-quality mesh extraction, we fine-tune the NeRF-based 3D VAE with an SDF-based representation~\cite{shen2023flexicubes}.
Considering that SDFs can be readily converged into meshes, this design choice broadens the utility of our model for more downstream tasks. 
}
To demonstrate the effectiveness of our extension, we conduct comprehensive experiments to evaluate the newly extended \nickname{} framework, including qualitative and quantitative assessments of image-conditioned 3D generation, along with comparisons to recent competitive baselines.

In summary, we contribute a 3D-representation-agnostic pipeline for building generic, high-quality 3D generative models. This pipeline provides opportunities to resolve a series of downstream 3D vision and graphics tasks. Specifically, we propose a novel 3D-aware reconstruction model that achieves high-quality 3D data encoding in an amortized manner. Learning in the compact latent space, our model demonstrates state-of-the-art 3D generation performance on the ShapeNet benchmark~\cite{shapenet2015}, surpassing both GAN-based and 3D diffusion-based approaches. Our method shows superior performance in monocular 3D reconstruction and conditional 3D generation on ShapeNet, FFHQ, and Objaverse datasets, with a fast inference speed, \ie, $3\times$ faster against existing latent-free 3D diffusion methods~\cite{anciukevivcius2023renderdiffusion}.



\section{Related Work}
\label{sec:related-works}

\heading{3D-aware GANs}
GANs~\cite{Goodfellow2014GenerativeAN} have shown promising results in generating photorealistic images~\cite{karras2019style,Brock2019LargeSG,karras_analyzing_2020}, inspiring researchers to explore 3D-aware generation~\cite{NguyenPhuoc2019HoloGANUL,platogan,pan_2d_2020}. Motivated by the recent success of neural rendering~\cite{park2019deepsdf,Mescheder2019OccupancyNetwork,mildenhall2020nerf}, researchers have introduced 3D inductive bias into the generation task~\cite{Chan2021piGANPI,Schwarz2020NEURIPS}, demonstrating impressive 3D-aware synthesis through hybrid designs~\cite{niemeyer2021giraffe,orel2021stylesdf,Chan2021EG3D,gu2021stylenerf,eva3d}. This has made 3D-aware generation applicable to a series of downstream applications~\cite{lan2022ddf_ijcv,zhang2023deformtoon3d,sun2022ide,sun2021fenerf}.
However, GAN-based methods suffer from mode collapse~\cite{ThanhTung2020CatastrophicFA} and struggle to model datasets with larger scale and diversity~\cite{dhariwal2021diffusion}. Besides, 3D reconstruction and editing with GANs require elaborately designed inversion algorithms~\cite{lan2022e3dge}.

\cm{
\heading{3D Generation via 2D Diffusion Models}
The remarkable success of 2D diffusion models~\cite{song2021scorebased,Jo2022DDPM} has motivated their adaptation for 3D content generation. Early approaches, such as score distillation sampling~\cite{poole2022dreamfusion,wang2023prolificdreamer}, attempt to distill 3D structures from 2D diffusion models. However, these methods face significant challenges, including computationally expensive optimization, mode collapse, and the Janus problem.
Some methods propose learning the 3D prior in a 2D manner~\cite{chan2023genvs,tewari2023forwarddiffusion,liu2023zero1to3,long2023wonder3d}. While these can produce photorealistic results, they lack view consistency and fail to fully capture the 3D structure.

Recent advancements have shifted toward a two-stage pipeline: generating multi-view images~\cite{shi2023MVDream,long2023wonder3d,shi2023zero123plus}, followed by feed-forward 3D reconstruction~\cite{hong2023lrm,xu2024instantmesh,tang2024lgm}. While these approaches have demonstrated promising results, their performance is inherently limited by the quality of multi-view image generation. Issues such as inconsistent viewpoints~\cite{liu2023zero1to3} and difficulties in scaling to higher resolutions~\cite{shi2023zero123plus} often hinder the overall effectiveness.
Furthermore, the reliance on a two-stage pipeline restricts 3D editing capabilities due to the absence of a unified, 3D-aware latent space, posing challenges for flexible and intuitive manipulation of 3D content.
To exploit rich pretrained 2D priors while ensuring geometric consistency, ~\cite{meng2025zero, lin2025diffsplat} synthesize splatter-based 3D representations directly with 2D diffusion models, thereby achieving 3D-consistent generation in a single 2D diffusion stage.
}

\heading{Native 3D Diffusion Models}
Recently, native 3D diffusion models~\cite{3dshape2vecset,zeng2022lion,zhang2024clay,lan2024ln3diff,li2024craftsman,lan2023gaussian3diff,zhang2024gaussiancube,Xiong_2025_SGP,direct3d,li2023diffusionsdf,chou2022diffusionsdf,chen2025sar3d} have been proposed to enable high-quality, efficient, and scalable 3D generation. These models follow a two-stage training process: first, encoding 3D objects into a latent space, and second, applying a latent diffusion model to the encoded latent codes.

While this pipeline is conceptually straightforward, existing methods vary in their choice of VAE input formats, latent space structures, and output 3D representations. Previous 3D diffusion pipeline adopts multi-view supervised auto-decoder as the stage-1 encoding tool~\cite{wang2023rodin,muller2023diffrf,Shue20223DNF,Dupont2022FromDT,lan2023gaussian3diff,Jun2023ShapEGC,zhang2024gaussiancube}. Then, the encoded 3D latent codes serve as the training corpus for diffusion. However, the auto-decoding stage leads to an unclean latent space and limited scalability~\cite{xu2023dmv3d}. Moreover, large latent codes, \eg, $256\times256\times96$~\cite{wang2023rodin}, hinder efficient diffusion learning~\cite{Hoogeboom2023simpleDE}.

Another line of work incorporates rendering operation into 3D diffusion training. RenderDiffusion~\cite{anciukevivcius2023renderdiffusion} and DMV3D~\cite{xu2023dmv3d} propose latent-free 3D diffusion by integrating rendering into diffusion sampling. However, this approach involves time-consuming volume rendering at each denoising step, significantly slowing down sampling.
SSDNeRF~\cite{ssdnerf} suggests a joint 3D reconstruction and diffusion approach, but requires a complex training schedule and shows performance only in single-category unconditional generation.
GaussianCube~\cite{zhang2024gaussiancube} proposes an optimal transport-driven 3DGS~\cite{kerbl3Dgaussians} encoding pipeline and leverages 3D-UNet for scalable 3D generation. However, it still leverages costly per-instance fitting before 3D diffusion training.
\cmfinal{
3DShape2Vec~\cite{3dshape2vecset} proposed the first vecset-based latent 3D generative model, and has been demonstrated scalable by recent industry efforts~\cite{direct3d}. 
However, these methods focus on 3D shape modeling and cannot generate colorful 3D mesh in a single stage.
OctFusion~\cite{Xiong_2025_SGP} presents a native 3D diffusion framework leveraging the compact octree latent representation, achieving both high-quality and efficient 3D generation. Nonetheless, its experiments are restricted to toy ShapeNet datasets, in contrast to our method, which enables category-free object generation.
Likewise, prior SDF-based 3D generative models~\cite{li2023diffusionsdf,chou2022diffusionsdf} suffer from the same limitation, with no evidence of general 3D object generation.
}
In contrast, our proposed \nickname{} trains 3D diffusion in a compressed VAE~\cite{Kingma2013AutoEncodingVB,Kosiorek2021NeRFVAEAG} latent space without rendering operations. As shown in Section~\ref{sec:experiment}, our method outperforms others in 3D generation and monocular 3D reconstruction, achieving three times faster speed. Additionally, we demonstrate conditional 3D generation over diverse datasets, whereas RenderDiffusion and SSDNeRF focus on simpler classes.
Other approaches, like 3DGen~\cite{gupta20233dgentriplanelatentdiffusion} and VolumeDiffusion~\cite{tang2023volumediffusion}, perform diffusion in the 3D latent space but heavily rely on 3D data (\eg, point clouds and voxels) and do not support monocular datasets like FFHQ~\cite{karras2018progressive}. Moreover, their methods are designed for U-Net, whereas our DiT-based architecture offers greater scalability.

\heading{Generalized 3D Reconstruction and View Synthesis}
To bypass the per-scene optimization of NeRF, researchers have proposed learning a prior model through image-based rendering~\cite{Wang2021IBRNetLM,yu2021pixelnerf,srt22,hong2023lrm,chen2024comboverse}. However, these methods are primarily designed for view synthesis and lack explicit 3D representations.
LoLNeRF~\cite{rebain2022lolnerf} learns a prior through auto-decoding but is limited to simple, category-specific settings. Moreover, these methods are intended for view synthesis and cannot generate new 3D objects.
VQ3D~\cite{Sargent2023VQ3DLA} adapts the generalized reconstruction pipeline to 3D generative models. However, it uses a 2D architecture with autoregressive modeling over a 1D latent space, ignoring much of the inherent 3D structure.
NeRF-VAE~\cite{Kosiorek2021NeRFVAEAG} directly models 3D likelihood with a VAE posterior but is constrained to simple 3D scenes due to the limited capacity of VAE.
\cm{Concurrently, LRM-line of work~\cite{hong2023lrm,tang2024lgm,wang2024crm} have proposed a feed-forward framework for generalized monocular reconstruction. However, they are still regression-based models and lack the latent space designed for generative modeling and 3D editing. Besides, they are limited to 3D reconstruction only and fail to support other modalities.
}

\begin{figure*}[t]
  \centering
  \includegraphics[width=1.0\textwidth]{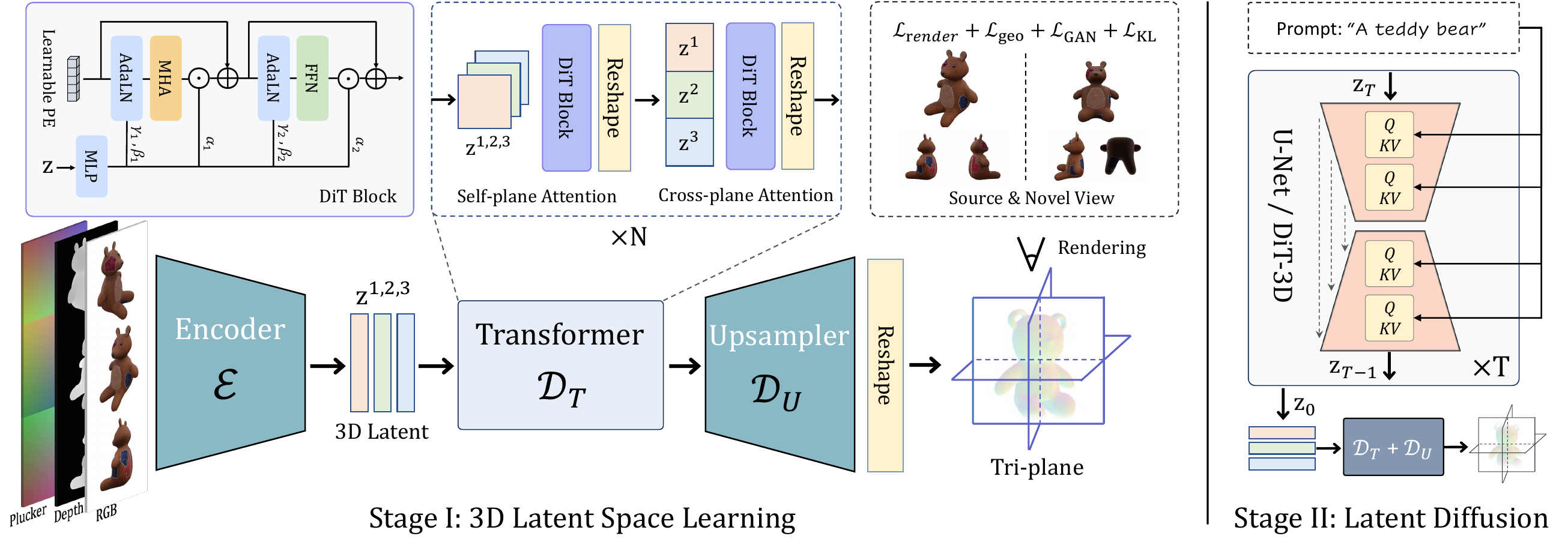} 
  \caption{\textbf{Pipeline of \nickname{}.} 
  In the 3D latent space learning stage, a convolutional encoder $\ldmencoder$ encodes a set of images $\set{I}$ into the KL-regularized latent space. The encoded 3D latent is further decoded by a 3D-aware DiT transformer $\mathcal{D}_T$, in which we perform self-plane attention and cross-plane attention.
  The transformer-decoded latent is up-sampled by a convolutional upsampler $\mathcal{D}_U$ towards a high-res tri-plane for rendering supervisions.
  In the next stage, we perform conditional diffusion learning over the compact latent space using either U-Net or DiT. The detailed architecture of DiT is shown in Fig.~\ref{fig:dit}.
  }
  \label{fig:overview}
\end{figure*}
\section{Scalable Latent Neural Fields Diffusion}
This section introduces our latent 3D diffusion model, which learns efficient diffusion prior over the compressed latent space by a dedicated variational autoencoder.
Specifically, the training goal is to learn a variational
encoder $\ldmencoder$ that maps a set of posed 2D image(s) $\set{I} = \{\image_{i}, ..., \image_{V}\}$,
of a 3D object to a latent code $\bz$, a denoiser $\model(\bz_t, t)$ to denoise the noisy latent code $\bz_t$ given diffusion time step $t$, 
and a decoder $\ldmdecoder$ (including a Transformer $\mathcal{D}_T$ and an Upsampler $\mathcal{D}_U$) to map $\bz_0$ to the 3D tri-plane $\widetilde{\mathcal{X}}$ corresponding to the input object.

Such a design offers several advantages: 1) By explicitly separating the 3D data compression and diffusion stage, we avoid the representation-specific 3D diffusion design~\cite{Zhou2021PVD,Jun2023ShapEGC,muller2023diffrf,Shue20223DNF,anciukevivcius2023renderdiffusion} and achieve a 3D representation/rendering-agnostic diffusion, which can be applied to any neural rendering techniques.
2) By leaving the high-dimensional 3D space, we reuse the well-studied Latent Diffusion Model (LDM) architecture~\cite{rombach2022LDM,vahdat2021score,Peebles2022DiT} for computationally efficient learning and achieve better sampling performance with faster speed.
3) The trained 3D compression model in the first stage serves as an efficient and general-purpose 3D tokenizer, whose latent space can be easily reused over downstream applications or extended to new datasets~\cite{yu2023mvimgnet,objaverseXL}.

In the following {subsections}, we first discuss the compressive stage with a detailed framework design in Sec.~\ref{sec:method:3d_latent_compression}.
Based on that, we introduce the 3D diffusion generation stage in Sec.~\ref{sec:method:3d_latent_diffusion} and present the condition injection in Sec.~\ref{sec:method:conditioning}.
The method overview is shown in Fig.~\ref{fig:overview}.

\subsection{Perceptual 3D Latent Compression}
\label{sec:method:3d_latent_compression}
As analyzed in Sec.~\ref{sec:intro}, directly leveraging neural fields for diffusion training hinders model scalability and performance.
Inspired by previous work~\cite{esser2020taming,rombach2022LDM}, we propose to {take multi-view image(s) as a proxy of the underlying 3D scene and} compress the input image(s) into a compact 3D latent space.
Though this paradigm is well-adopted in the image domain~\cite{esser2020taming,rombach2022LDM} with similar trials in specific 3D tasks~\cite{lan2022e3dge,Mi2022im2nerf,cai2022pix2nerf,Sargent2023VQ3DLA},
we, for the first time, demonstrate that a high-quality compression model is feasible, whose latent space serves as a compact proxy for efficient diffusion learning.

\heading{Encoder}
Given a set of image(s) $\set{I}$ of an 3D object where each image within the set $\image \in \real^{H \times W \times 3}$ is an observation of an underlying 3D object from viewpoints $\set{C} = \{{\rvc}_{1}, \dots, {\rvc}_{V} \}$, \nickname{} adopts a convolutional encoder $\ldmencoder$ to encode the image set $\set{I}$ into a latent representation $\bz \sim \ldmencoder(\set{I})$.
To inject camera condition, we concatenate Plucker coordinates $\rvr_i = (\rvd_i, \rvp_i \times \rvd_i) \in \real^{6}$ channel-wise as part of the input~\cite{sitzmann2021lfns}, where $\rvd_i$ is the normalized ray direction, $\rvp_i$ is the camera origin corresponding to the camera $\rvc_i$, and $\times$ denotes the cross product.
For challenging datasets like Objaverse~\cite{objaverse}, we also concatenate the rendered depth map, making our input a dense 3D colored point cloud~\cite{wu2023multiview}.

Unlike existing works~\cite{Sargent2023VQ3DLA,esser2020taming} that operate on 1D order latent and ignore the internal structure, we choose to output 3D latent $\bz \in \real^{h \times w \times d \times c}$ to facilitate 3D-aware operations,
 where $h=H/f, w=W/f$ are the spatial resolution with down-sample factor $f$, and $d$ denotes the 3D dimension.
Here we set $f=8$ and $d=3$ to make $\bz \in \real^{h \times w \times 3 \times c}
$ a tri-latent, which is similar to tri-plane~\cite{Chan2021EG3D,Peng2020ConvolutionalON} but in the compact 3D latent space.
We further impose \emph{KL-reg}~\cite{Kingma2013AutoEncodingVB} to encourage a well-structured latent space to facilitate diffusion training~\cite{rombach2022LDM,vahdat2021score}.

\heading{Decoder Transformer}
The decoder aims to decode the compact 3D codes $\bz$ for high-quality 3D reconstruction.
Existing image-to-3D methods~\cite{cai2022pix2nerf,Chan2021EG3D,gu2023control3diff,lan2022e3dge} adopt convolution as the building block, which lacks 3D-aware operations and impedes information flow in the 3D space.
Here, we adopt ViT~\cite{Peebles2022DiT,dosovitskiy2020vit} as the decoder backbone due to its flexibility and effectiveness.
Inspired by Rodin~\cite{wang2023rodin}, we made the following reformulation to the raw ViT decoder to encourage 3D inductive bias and avoid the mixing of uncorrelated 3D features:
{1)} \textit{Self-plane Attention Block}. 
Given the input $\bz \in \real^{l \times 3 \times c}$ where $l = h \times w$ is the sequence length, we treat each of the three latent planes as a data point and conduct self-attention within itself.
This operation is efficient and encourages local feature aggregation.
{2)} \textit{Cross-plane Attention Block}.
To further encourage 3D inductive bias, we roll out $\bz$ as a long sequence ${l \times 3 \times c} \rightarrow {3l \times c}$ to conduct attention across planes,
so that all tokens in the latent space could attend to each other.
In this way, we encourage global information flow for more coherent 3D reconstruction and generation.
Compared to Rodin, our design is fully attention-based and naturally supports parallel computing without the expensive axis pooling aggregation.

Empirically, we observe that using DiT~\cite{Peebles2022DiT} block and injecting the latent $\bz$ as conditions yields better performance compared to the ViT~\cite{dosovitskiy2020vit,oquab2023dinov2} block, which takes the latent $\bz_0$ as the regular input. 
Specifically, the adaptive layer norm (adaLN) layer~\cite{Peebles2022DiT} fuses the input latent $\bz$ with the learnable positional encoding for attention operations. 
Moreover, we interleave the two types of attention layers to make sure the overall parameters count consistent with the pre-defined DiT length, ensuring efficient training and inference.
As all operations are defined in the token space, the decoder achieves efficient computation against Rodin~\cite{wang2023rodin} while promoting 3D priors.

\heading{Decoder Upsampler}
After all the attention operations, we obtain the tokens from the last transformer layer $\widetilde{\bz}$ as the output.
The context-rich tokens are reshaped back into spatial domain~\cite{He2021MaskedAA} and up-sampled by a convolutional decoder to the final tri-plane representation with shape $\hat{H} \times \hat{W} \times3C$. 
Here, we adopt a lighter version of the convolutional decoder for efficient upsampling, where the three spatial latent of $\widetilde{\bz}$ are processed in parallel.

\heading{Learning a Perceptually Rich and Intact 3D Latent Space}
Adversarial learning~\cite{Goodfellow2014GenerativeAN} has been widely applied in learning a compact and perceptually rich latent space~\cite{esser2020taming,rombach2022LDM}. 
In the 3D domain, the adversarial loss can also encourage correct 3D geometry when novel-view reconstruction supervisions are inapplicable~\cite{Chan2021piGANPI,Sargent2023VQ3DLA,kato2019vpl}, \eg, the monocular dataset such as FFHQ.
Inspired by previous research~\cite{Sargent2023VQ3DLA,kato2019vpl}, we leverage adversarial loss to bypass this issue.
Specifically, we impose an input-view discriminator to maintain perceptually-reasonable input view reconstruction, and an auxiliary novel-view discriminator to distinguish the rendered images between the input and novel views.
We observe that if asking the novel-view discriminator to differentiate novel-view renderings and real images instead, the reconstruction model will suffer from \emph{posterior collapse}~\cite{Lucas2019UnderstandingPC}, which outputs input-irrelevant but high-fidelity results to fool the novel-view discriminator.
This phenomenon has also been observed by Kato \etal~\cite{kato2019vpl}. 

\heading{Training}
After the decoder $\ldmdecoder$ decodes a high-resolution neural field $\hat{\bz_0}$ from the latent, we have $\hat{\image} = \diffrender(\widetilde{\mathcal{X}}) = \diffrender(\ldmdecoder(\bz)) = \diffrender(\ldmdecoder(\ldmencoder(\set{\image})))$, where $\diffrender$ stands for differentiable rendering~\cite{Tewari2021AdvancesIN} and we take $\ldmdecoder(\bz)=\diffrender(\ldmdecoder(\bz))$ for brevity.
Here, we choose $\widetilde{\mathcal{X}}$ as tri-plane~\cite{Chan2021EG3D,Peng2020ConvolutionalON} and $\diffrender$ as volume rendering~\cite{mildenhall2020nerf} for experiments. 
Note that our compression model is 3D representations/rendering agnostic and new neural rendering techniques~\cite{kerbl3Dgaussians} can be easily integrated by alternating the decoder architecture~\cite{szymanowicz23splatter}.
The final training objective reads as
\begin{align}
    \label{eq:stage1_loss}
    \loss({\bphi,\bpsi}) = \loss_\text{render} + \lambda_\text{geo}\loss_\text{geo} + \lambda_\text{kl}\loss_\text{KL} + \lambda_\text{GAN}\loss_\text{GAN},
\end{align}

where $\loss_\text{render}$ is a mixture of $L_1$ and perceptual loss~\cite{zhang2018perceptual}, $\loss_\text{reg}$ encourages smooth geometry~\cite{weng_personnerf}, $\loss_\text{KL}$ is the loss \emph{KL-reg} to regularize a structured latent space~\cite{rombach2022LDM}, and $\loss_\text{GAN}$ improves perceptual quality and enforces correct geometry for monocular datasets. 
Note that $\loss_\text{render}$ is applied to both input-view and randomly sampled novel-view images.

\cm{
To facilitate 3D mesh extraction, we further finetune the model to the hybrid representation Flexicubes~\cite{shen2023flexicubes,xu2024instantmesh} with the extra $\mathcal{L}_\text{flex}$ loss:
\begin{equation}
    \begin{aligned}
        \mathcal{L}_\text{flex} &= \lambda_\text{normal}\mathcal{L}_\text{normal} + \lambda_\text{reg}\mathcal{L}_\text{reg}, 
    \end{aligned}
\end{equation}
where $\mathcal{L}_\text{normal}$ is MAE loss between rendered normal and ground truth, $\mathcal{L}_\text{reg}$ is regularization term for Flexicubes parameters~\cite{shen2023flexicubes}.  The corresponding loss weights are represented by $\lambda_\text{*}$.
Similar to LATTE3D~\cite{xie2024latte3d}, we only fine-tune the decoder of the 3D VAE at this stage for stabilized training.
}

For category-specific datasets such as ShapeNet~\cite{shapenet2015}, we only supervise \emph{one} novel view, which already yields good enough performance.
For category-free datasets with diverse shape variations, \eg, Objaverse~\cite{objaverse}, we supervise \emph{four} novel views.
Our method is more data-efficient against the existing state-of-the-art 3D diffusion method~\cite{ssdnerf,muller2023diffrf}, which requires $50$ views to converge.
The implementation details are included in the supplementary material.

\subsection{Latent Diffusion and Denoising}
\label{sec:method:3d_latent_diffusion}

\begin{figure}[t]
  \centering
  \includegraphics[width=0.75\linewidth]{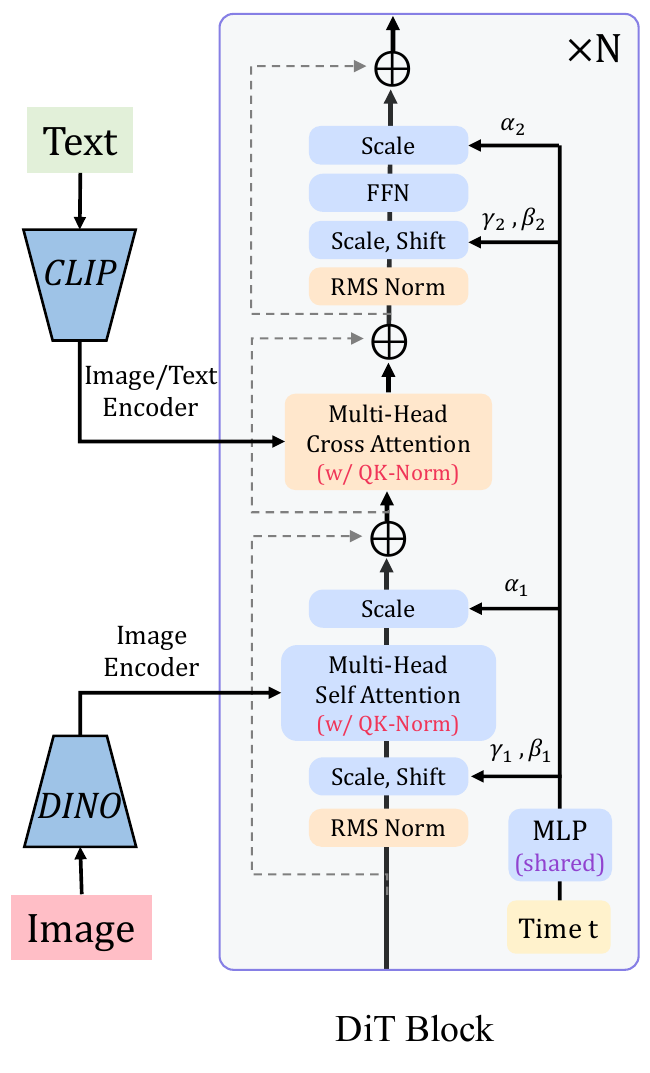} 
  \vspace{-2mm}
  \caption{\textbf{Diffusion training of \nickname{}.} 
  \cm{We adopt DiT architecture with AdaLN-single~\cite{chen2023pixartalpha} and QK-Norm~\cite{megavit,esser2020taming}. For both conditioning modalities, we incorporate the conditional features using attention mechanisms. Specifically, for CLIP-based conditioning, we employ cross-attention blocks to inject the condition, following the approach used in PixelArt~\cite{chen2023pixartalpha}. For image-conditioned 3D generation, we additionally concatenate DINO patch features into the self-attention block.}
  }
  \label{fig:dit}
\end{figure}

\heading{Latent Diffusion Models}
LDM~\cite{rombach2022LDM,vahdat2021score} is designed to acquire a prior distribution $p_\btheta(\bz_0)$ within the perceptual latent space, whose training data is the latent obtained online from the trained $\ldmencoder$.
Here, we use the score-based latent diffusion model~\cite{vahdat2021score}, which is the continuous derivation of DDPM variational objective~\cite{Jo2022DDPM}.
Specifically, the denoiser $\model$ parameterizes the score function score~\cite{song2021scorebased} as $\nabla_{\bz_t} \log p(\bz_t) := - \modelwithinp/{\sigma_t}$, with continuous time sequence $t$.
By training to predict a denoised variant of the noisy input $\bz_t$, $\model$ gradually learns to denoise from a standard Normal prior $\N(\bzero, \beye)$ by solving a reverse SDE~\cite{Jo2022DDPM}.
Following LSGM~\cite{vahdat2021score}, we formulate the learned prior at time $t$ geometric mixing $\beps_\theta(\bz_t, t) := \sigma_t (1 - \balpha) \odot \bz_t + \balpha \odot \beps'_\theta(\bz_t, t)$, where  $\beps'_\theta(\bz_t, t)$ is the denoiser output and $\alpha \in [0, 1]$ is a learnable scalar coefficient.
Intuitively, this formulation can bring the denoiser input closer to a standard Normal distribution, on which the reverse SDE can be solved faster.
Similarly, Stable Diffusion~\cite{rombach2022LDM,sdxl} also scales the input latent by a factor to maintain a unit variance, which is pre-calculated on the billion-level dataset~\cite{Schuhmann2022LAION5BAO}.
The training objective reads as
%
\begin{align}
\lsimplediff = \expec_{\ldmencoder(\image), \beps \sim \mathcal{N}(0, 1),  t}\Big[\dfrac{w_t}{2} \Vert \beps - \model(\bz_{t},t, c) \Vert_{2}^{2}\Big] \,,
\label{eq:ldmloss}
\end{align}
%
where $t\sim \U[0, 1]$ and $w_t$ is an empirical time-dependent weighting function, $c$ is the corresponding condition. 

The denoiser $\model$ is realized by a time-dependent U-Net~\cite{ronneberger_u-net_2015}, as visualized in Fig.~\ref{fig:overview}.
During training, we obtain $\bz_{0}$ online from the fixed $\ldmencoder$, roll-out the tri-latent $h \times w \times 3 \times c \rightarrow h \times (3w) \times c$, and add time-dependent noise to get $\bz_{t}$.
Here, we choose the importance sampling schedule~\cite{vahdat2021score} with $velocity$~\cite{Meng2022OnDO} parameterization, which yields more stable behavior against $\epsilon$ parameterization for diffusion learning.

\cm{
\heading{3D Flow Matching}
Beyond the vanilla LSGM framework, we also explore the flow matching~\cite{liu2022flow,albergo2023stochastic,lipman2022}-based diffusion framework. Specifically, flow matching involves training a neural network $\model$ to predict the {velocity} $v$ of the noisy input $\bz_t$ with the straight-line trajectory. After training, $\model$ can sample from a standard Normal prior $\gN(0, I)$ by solving the reverse ODE/SDE~\cite{Karras2022edm}.
In our case, the training data point is the compact tri-plane latent code. Note that compared to U-Net architecture that adopts roll-out tri-latent, in flow matching training, we opt for DiT~\cite{Peebles2022DiT} with full attention~\cite{esser2024sd3} over $L = h \times w \times 3$, as detailed in Fig.~\ref{fig:dit}.
The training objective now reads as

\begin{equation}
  \mathcal{L}_\text{FM} = -\frac{1}{2} \mathbb{E}_{\ldmencoder(\image), \epsilon\sim \mathcal{N}(0, I), t}
  \left[ w_t^\text{FM} \lambda_t' \Vert \epsilon - \model(
  \bz_{t}, t, c)
  \Vert_{2}^2 \right]\;,
  \label{eq:flowmatching}
\end{equation}

where $\lambda_t:= \log \frac{a_t^2}{b_t^2}$ denotes \emph{signal-to-noise ratio}, and $\lambda_t'$ denotes its derivative. By setting $w_t = \frac{t}{1-t}$ with $z_t = (1-t) x_0 + t \epsilon$, flow matching defines the forward process as a straight path between the data distribution and the Normal distribution. \cm{The network $\model$ directly predicts the \emph{velocity} $v_\Theta$, and please check the Sec.2 of SD-3~\cite{esser2024sd3} for theoretical derivation.}
}

After training, the denoised samples will be decoded to the 3D neural field (\ie, tri-plane here) with a single forward pass through $\ldmdecoder$, on which neural rendering can be applied.

\subsection{Conditioning Mechanisms}
\label{sec:method:conditioning}
Compared to LRM~\cite{hong2023lrm} line of work, which \cm{intrinsically relies on image(s) as the input}, our native diffusion-based method enables more flexible 3D generation from diverse conditions.
As shown in Fig.~\ref{fig:overview} and Fig.~\ref{fig:dit}, we propose to inject CLIP embeddings~\cite{Radford2021CLIP} and DINO embeddings~\cite{oquab2023dinov2} into the latent 3D diffusion model to support image/text-conditioned 3D generation.
Given the input condition $\by$, the diffusion model formulates the conditional distribution $p(\bz | \by)$ with $\model(\bz_t, t, \by)$.
The inputs $\by$ can be text captions for datasets like Objaverse, or images for general 3D datasets like ShapeNet, FFHQ, and Objaverse. 

\heading{Text Conditioning}
For datasets with text captions, we follow Stable Diffusion~\cite{rombach2022LDM} to directly leverage the CLIP text encoder $\text{CLIP}_{T}$ to encode the text caption as conditions. 
All output tokens $77 \times 768$ are used and injected into the diffusion denoiser with cross attention blocks. \cm{For both U-Net architecture and DiT architecture, the CLIP text conditions are injected via cross attention.}

\begin{figure*}[t!]
\centering
  \includegraphics[width=0.9\textwidth]{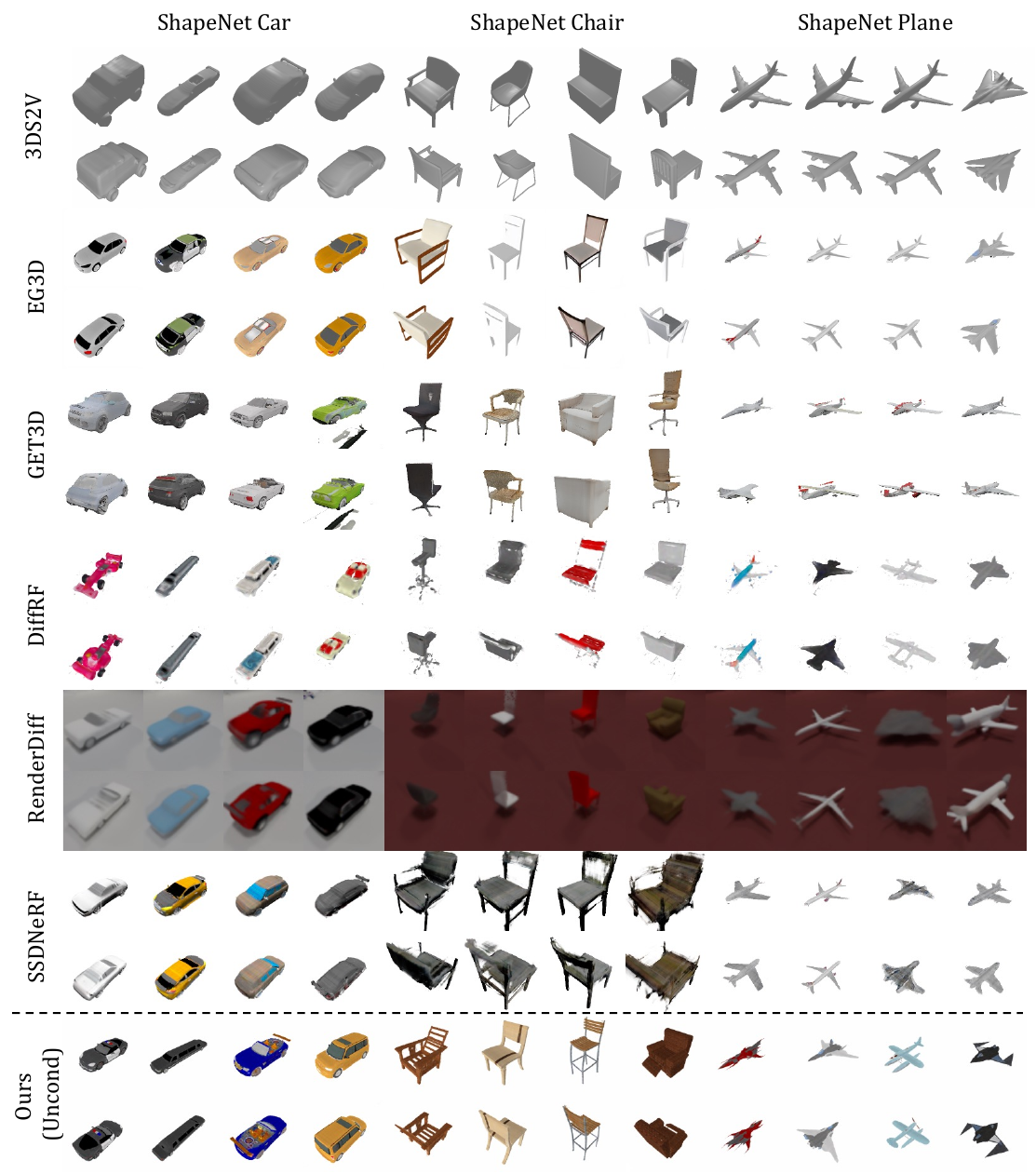} 
  \caption{\textbf{ShapeNet Unconditional Generation.} We show four samples for each method. Zoom in for the best view.}
  \label{fig:shapenet:uncond}
\end{figure*}
\heading{Image Conditioning}
\cm{To support more flexible 3D content creation, \nickname{} further supports image conditions.}. Specifically, \cm{for category-specific 3D dataset}, we first encode the input $\image$ corresponding to the latent code $\bz_0$ using the CLIP image encoder $\text{CLIP}_{I}$ and adopt the output embedding as the condition.
To support both image and text conditions, we re-scale the image latent code with a pre-calculated factor to match the scale of the text latent. Cross attention is also leveraged to inject CLIP image features.

\cm{For model trained on category-free 3D dataset like Objaverse~\cite{objaverse}, we further incorporate DINO~\cite{oquab2023dinov2} features to improve the reconstruction fidelity. Rather than introducing another cross-attention module, we incorporate the DINO features by pre-pending the patch tokens in the self-attention layers, similar to SD-3~\cite{esser2024sd3}. Compared to leveraging CLIP features for image conditions, introducing low-level DINO features improves the 3D generation faithfulness and fidelity in image-conditioned setting.}

\heading{Classifier-free Guidance}
We adopt \emph{classifier-free guidance}~\cite{Ho2022ClassifierFreeDG} for latent conditioning to support conditional and unconditional generation. 
During diffusion model training, we randomly zero out the corresponding conditioning latent embeddings with $15\%$ probability to jointly train unconditional and conditional settings.
During sampling, we perform a linear combination of the conditional and unconditional score estimates:
\begin{align}
    \hat{\beps}_\theta(
    \bz_t, \boldsymbol{\tau}_\theta(y)
    ) = s\model(\bz_t, \boldsymbol{\tau}_\theta(y)) + (1-s)\model(\bz_t)
    ,
    \label{eq:guidance}
\end{align}
where $s$ is the guidance scale to control the mixing strength to balance sampling diversity and quality. 

\cm{
\heading{Stable and Efficient Training}
For efficient and scalable training, we leverage 
BFloat16~\cite{bfloat16} with FlashAttention~\cite{dao2022flashattention,dao2023flashattention2} enabled.
All conditioning transformer blocks follow a pre-norm design~\cite{xiong2020layernormalizationtransformerarchitecture} with QK-Norm~\cite{megavit} enabled. RMSNorm~\cite{zhang19rmsnorm} is leveraged for efficient AdaLN operation defined in diffusion transformer, following SD-3~\cite{esser2024sd3}.
}

\section{Experiments} 
\label{sec:experiment}
\begin{figure*}[t]
  \includegraphics[width=1.0\textwidth]{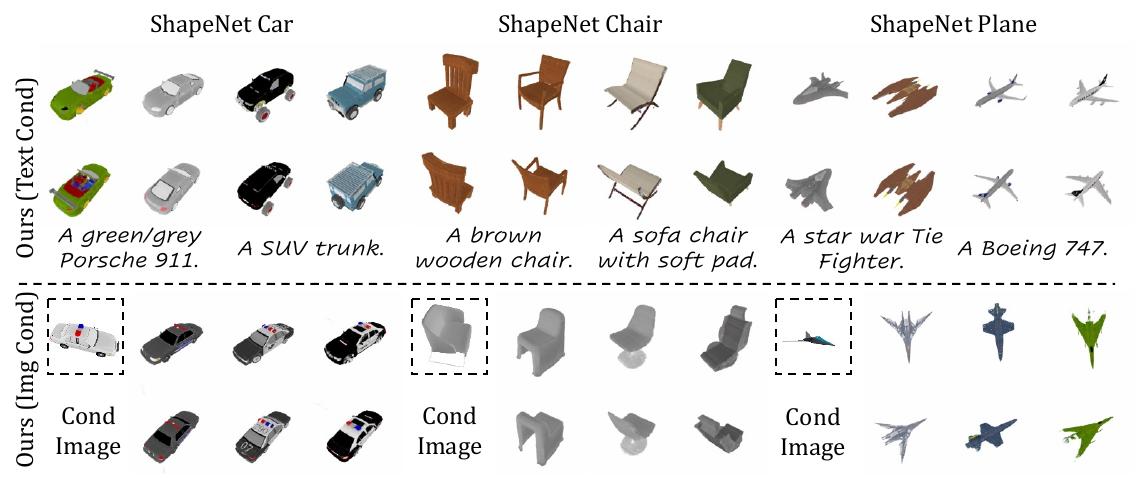} 
  \caption{\textbf{ShapeNet Conditional Generation.} We show conditional generation with both texts and image as inputs. Zoom in for the best view.}
  \label{fig:shapenet:cond}
\end{figure*}

\heading{Datasets} 
Following most previous work, we use ShapeNet~\cite{shapenet2015} to benchmark 3D generation performance. 
We use the Car, Chair, and Plane categories with 3514, 6700, and 4045 instances, respectively. 
Each instance is randomly rendered from $50$ views following 
a spherical uniform distribution.
Moreover, to evaluate the performance over diverse, high-quality 3D datasets, we also include experiments on Objaverse~\cite{objaverse}, which is the largest 3D dataset with challenging categories and complicated geometry. 
We use the renderings provided by G-Objaverse~\cite{qiu2023richdreamer} and choose a high-quality subset with around $176K$ 3D instances, where each consists of $40$ random views.
Text prompts from Cap3D~\cite{luo2023scalable} are used for text-conditioned 3D generation training.

\heading{Training Details}
\cm{For VAE training} with ShapeNet and FFHQ, we adopt a monocular input setting with $V=1$ and target rendering size $128 \times 128$. 
For Objaverse, we adopt $V=6$ posed RGB-D renderings with $256\times256$ resolution as inputs to guarantee a thorough coverage of the 3D object. The target rendering size is set to $192 \times 192$.
The encoder $\ldmencoder$ has down-sample factor $f=8$ and the decoder upsampler $\mathcal{D}_U$ outputs tri-plane with size $\hat{H}=\hat{W}=128$ and $C=32$.
To trade off rendering resolution and training batch size, we impose supervision over $80\times80$ randomly cropped patches.
For adversarial loss, we use DINOv2~\cite{oquab2023dinov2} in vision-aided GAN~\cite{kumari2021ensembling} with non-saturating GAN loss~\cite{Gulrajani2017wgangp} for discriminator training. 
\cm{For Flexicube fine-tuning, we directly render the whole image with resolution $256 \times 256$. We follow the same initialization strategy in InstantMesh~\cite{xu2024instantmesh}, where the weights of $\sigma$ prediction MLP are flipped to accommodate for SDF prediction.}

\cm{On the generation side, } for conditional diffusion training, we use CLIP image embeddings for ShapeNet and FFHQ, and CLIP text embeddings from the official text caption for Objaverse.
\cm{For image-conditioned training, we randomly select an image from the dataset corresponding to each 3D instance as the conditioning input.}
\cm{The DiT-based 3D generative model uses the DiT-L architecture~\cite{Peebles2022DiT}, which consists of $24$ transformer layers with $16$ attention heads and $1024$ latent dimensions.}
Both the autoencoding model and the diffusion model are trained for $800K$ iterations, which take around $7$ days with $8$ A100 GPUs in total.

\subsection{\cm{Metrics and Baselines}}
\heading{Evaluating Unconditional Generation}
For unconditional 3D generation, we adopt category-specific ShapeNet~\cite{shapenet2015} and adopt state-of-the-art GAN-based methods: EG3D~\cite{Chan2021EG3D}, GET3D~\cite{gao2022get3d} as well as recent diffusion-based 3D generative methods: DiffRF~\cite{muller2023diffrf}, RenderDiffusion~\cite{anciukevivcius2023renderdiffusion} and SSDNeRF~\cite{ssdnerf} as baselines.
\cmfinal{
We also include a canonical 3D shape generation method 3DShape2Vec~\cite{3dshape2vecset} for geometry qualitative comparison.
}
Since \nickname{} only leverages $V=2$ for ShapeNet experiments, for SSDNeRF, we include both the official $50$-views trained SSDNeRF$_\text{V=50}$ version as well as the reproduced SSDNeRF$_\text{V=3}$ for fair comparison. We find SSDNeRF fails to converge with $V=2$.
We set the guidance scale in Eq.~(\ref{eq:guidance}) to $s=0$ for unconditional generation, and $s=6.5$ for all conditional generation sampling.

Regarding the metrics, following prior work~\cite{ssdnerf}, we adopt both 2D and 3D metrics to benchmark the generation performance: Fréchet Inception Distance (FID@50K)~\cite{heusel_gans_2018} and Kernel Inception Distance (KID@50K)~\cite{binkowski2018demystifying} to evaluate 2D renderings, as well as Coverage Score (COV) and Minimum Matching Distance (MMD) to benchmark 3D geometry.
We compute all metrics at $128 \times 128$ resolution to ensure fair comparisons across all baselines.

\heading{Evaluating Text-to-3D Generation}
Regarding text-conditioned 3D generation methods, we compare against Point-E~\cite{nichol2022pointe}, Shape-E~\cite{Jun2023ShapEGC}, and 3DTopia~\cite{hong20243dtopia}. \cm{Moreover, we also include the comparison with the latest 3DGS-based text-to-3D generative model, GaussianCube V1.1~\cite{zhang2024gaussiancube} for reference. Note that GaussianCube V1.1 is trained on auxiliary data beyond Objaverse and adopts 3D captions from proprietary language model~\cite{OpenAI_GPT4_2023}, rather than Cap3D. The result is included here for reference, and we emphasize that the comparison is indeed unfair.}
CLIP score~\cite{Radford2021CLIP} is reported following the previous works~\cite{Jun2023ShapEGC}, \cm{with aesthetic scores MUSIQ-AVA~\cite{ke2021musiq} and Q-Align~\cite{wu2023qalign} also included}.

\heading{Evaluating Image-to-3D Generation}
Regarding image-conditioned 3D generation methods, we compare the proposed method with three lines of methods: \emph{single-image to 3D methods}: OpenLRM~\cite{openlrm,hong2023lrm}, Splatter Image~\cite{szymanowicz23splatter}, \emph{multi-view images to 3D methods}: One-2-3-45~\cite{liu2023one2345}, CRM~\cite{wang2024crm}, Lara~\cite{LaRa}, LGM~\cite{tang2024lgm}, and \emph{native 3D diffusion models}: Shape-E~\cite{Jun2023ShapEGC}.


\begin{table}[h!]
\caption{
\textbf{Quantitative Metrics on Text-to-3D.} 
The proposed method outperforms Point-E and Shape-E on CLIP scores over two different backbones.
}
\label{tab:t23d-clipscore}
\resizebox{1.0\linewidth}{!}{
\begin{tabular}{lcccc}
\toprule
Method & ViT-B/32$\uparrow$ & ViT-L/14$\uparrow$ & \cm{MUSIQ-AVA $\uparrow$} & \cm{Q-Align $\uparrow$} \\
\midrule
Point-E & 26.35 & 21.40 & \cm{4.08} & \cm{1.21} \\
Shape-E & 27.84 & 25.84 & \cm{3.69} & \cm{1.56} \\
3DTopia & 28.10 & 26.31 & \cm{3.31} & \cm{1.42} \\
GaussianCube-V1.1 & \textbf{29.84} & 27.36 & \cm{\textbf{4.89}} & \cm{\textbf{2.86}} \\
\midrule
Ours & {29.12} & \textbf{27.80} & {4.16} & {2.22} \\
\bottomrule
\end{tabular}
}
\end{table} 
\cm{
Quantitatively, we benchmark rendering metrics with CLIP-I~\cite{Radford2021CLIP},
FID~\cite{heusel_gans_2018}, KID~\cite{binkowski2018demystifying}, and MUSIQ-koniq~\cite{ke2021musiq}. For 3D quality metrics, we adopt Point cloud FID (P-FID), Point cloud KID (P-KID), Coverage Score (COV), and Minimum Matching Distance (MMD) as the metrics. Following previous works~\cite{nichol2022pointe,3dshape2vecset,yariv2023mosaicsdf}, we adopt the pre-trained PointNet++ provided by Point-E~\cite{nichol2022pointe} to calculate P-FID and K-FID.
Qualitatively, GSO~\cite{gso} dataset is used for visually inspecting image-conditioned generation. 
}

\begin{table*}[t]
    \centering
    \caption{\textbf{Quantitative Comparison of Unconditional Generation on ShapeNet}. The proposed \nickname{} shows satisfactory performance on single-category generation.
    }
    \resizebox{0.72\textwidth}{!}{ 
    \begin{tabular}{lccccc}
        \toprule        Category&Method&FID@50K$\downarrow$&KID@50K(\%)$\downarrow$&COV(\%)$\uparrow$&MMD(\textperthousand)$\downarrow$   \\
        \midrule
        \multirow{5}*{Car}&EG3D~\cite{Chan2021EG3D}&\cellcolor{orange}33.33&\cellcolor{orange}1.4&35.32& 3.95  \\
        &GET3D~\cite{gao2022get3d} &\cellcolor{yellow}41.41&\cellcolor{yellow}1.8&37.78&3.87  \\
        
        &DiffRF~\cite{muller2023diffrf}&75.09&5.1&29.01&4.52   \\
        &RenderDiffusion~\cite{anciukevivcius2023renderdiffusion}&46.5&4.1&-&-   \\
        &${\text{SSDNeRF}_\text{V=3}}$~\cite{ssdnerf}&47.72&2.8&\cellcolor{yellow}37.84&\cellcolor{yellow}3.46   \\
        &${\text{SSDNeRF}_\text{V=50}^{*}}$~\cite{ssdnerf}&45.37&2.1&\cellcolor{tablered}{67.82}&\cellcolor{orange}2.50   \\
        &\nickname{}~\textbf{(Ours)} &\cellcolor{tablered}{17.6}&\cellcolor{tablered}{0.49}&\cellcolor{orange}{43.12}& \cellcolor{tablered}{2.32}  \\
        \midrule
        \multirow{5}*{Plane}&EG3D~\cite{Chan2021EG3D}&\cellcolor{orange}14.47&\cellcolor{orange}0.54&18.12&4.50   \\
        &GET3D~\cite{gao2022get3d} &26.80&1.7&21.30&4.06  \\
        &DiffRF~\cite{muller2023diffrf}&101.79&6.5&\cellcolor{yellow}37.57&\cellcolor{yellow}3.99   \\
        &RenderDiffusion~\cite{anciukevivcius2023renderdiffusion}&43.5&5.9&-&-   \\
        &${\text{SSDNeRF}_\text{V=3}}$~\cite{ssdnerf}&\cellcolor{yellow}21.01&\cellcolor{yellow}1.0&\cellcolor{orange}42.50&\cellcolor{orange}2.94   \\
        
        &\nickname{}~\textbf{(Ours)} &\cellcolor{tablered}{8.84}&\cellcolor{tablered}{0.36}&\cellcolor{tablered}{43.40}&\cellcolor{tablered}{2.71}  \\
        \midrule
        
        \multirow{5}*{Chair}&EG3D~\cite{Chan2021EG3D}&\cellcolor{orange}26.09&\cellcolor{orange}1.1&19.17&10.31   \\
        &GET3D~\cite{gao2022get3d} &\cellcolor{yellow}35.33&\cellcolor{yellow}1.5&\cellcolor{yellow}28.07& \cellcolor{yellow}9.10  \\
        &DiffRF~\cite{muller2023diffrf}&99.37&4.9&17.05&14.97   \\
        &RenderDiffusion~\cite{anciukevivcius2023renderdiffusion}&53.3&6.4&-&-   \\
        &${\text{SSDNeRF}_\text{V=3}}$~\cite{ssdnerf}&65.04&3.0&\cellcolor{tablered}{47.54}&\cellcolor{orange}6.71   \\
        
        &\nickname{}~\textbf{(Ours)} &\cellcolor{tablered}{16.9}&\cellcolor{tablered}{0.47}&\cellcolor{orange}{47.1}&\cellcolor{tablered}{5.28}  \\
        \bottomrule
    \end{tabular}}
    \label{tab:shapenet}%
\end{table*}%
\subsection{\cm{Evaluation}}
\heading{Unconditional Generation on ShapeNet}
To evaluate our methods against existing 3D generation methods, we include the quantitative and qualitative results for unconditional single-category 3D generation on ShapeNet in Tab.~\ref{tab:shapenet} and Fig.~\ref{fig:shapenet:uncond}.
We evaluate all baseline 3D diffusion methods with $250$ DDIM steps and GAN-based baselines with $psi=0.8$ to guarantee each sample is intact for COV/MMD evaluation.
For FID/KID evaluation, we re-train the baselines and calculate the metrics using a fixed upper-sphere ellipsoid camera trajectory~\cite{sitzmann2019srns} across all datasets.
For COV/MMD evaluation, we randomly sample $4096$ points around the extracted sampled mesh and ground truth mesh surface and adopt Chamfer Distance for evaluation.

As shown in Tab.~\ref{tab:shapenet}, \nickname{} achieves quantitatively better performance against all GAN-based baselines regarding rendering quality and 3D coverage.
Fig.~\ref{fig:shapenet:uncond} further demonstrates that GAN-based methods suffer greatly from mode collapse: in the ShapeNet Plane category, both EG3D and GET3D are limited to the white civil airplanes, which is fairly common in the dataset. Our methods can sample more diverse results with high-fidelity texture.

Compared to diffusion-based baselines, \nickname{} delivers higher visual quality and stronger quantitative performance..
SSDNeRF$_{\text{V=50}}$ shows a better coverage score, which benefits from leveraging more views during training.
However, our method with $V=2$ shows comparative performance against SSDNeRF$_\text{V=3}$ on the ShapeNet Chair and even better performance on the remaining datasets. 

\begin{figure*}[b]
  \includegraphics[width=1.0\textwidth]{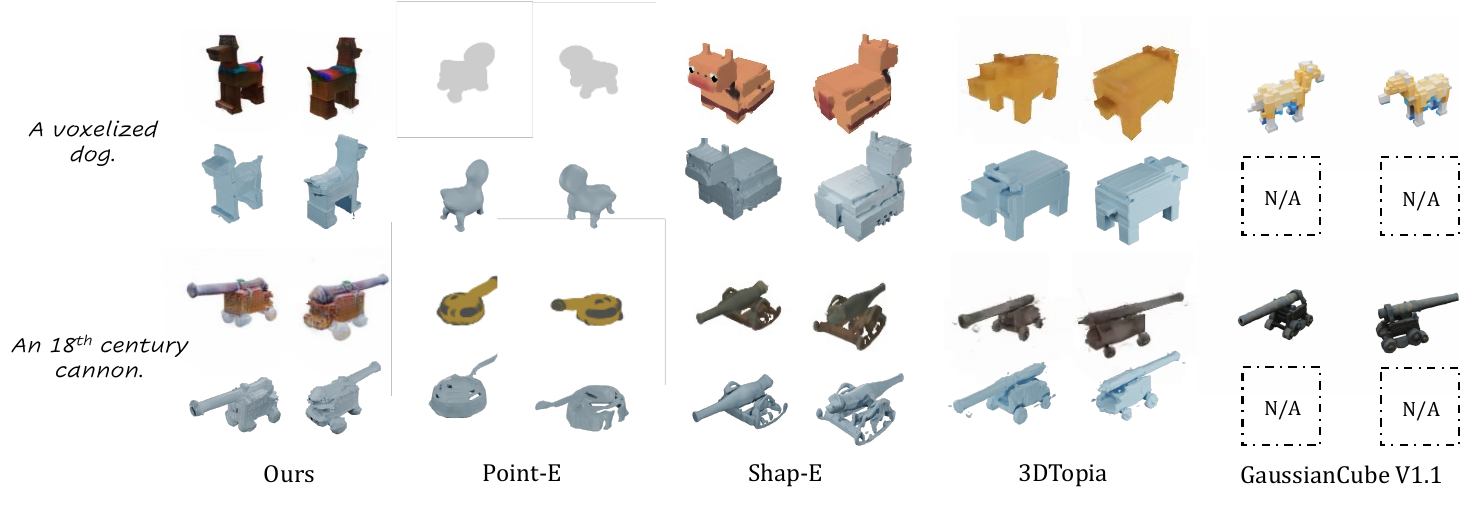} 
  \caption{\textbf{Qualitative Comparison of Text-to-3D} We showcase uncurated samples generated by \nickname{} on ShapeNet three categories. We visualize two views for each sample. Better zoom in.}
  \label{fig:t23d:oai-comparison}
\end{figure*}
\heading{\cm{Text-to-3D Generation}}
Conditional 3D generation has the potential to streamline the 3D modeling process in both the gaming and film industries. 
As visualized in Fig.~\ref{fig:shapenet:cond}, we present our conditional generation performance on the ShapeNet dataset, where either text or image serves as the input prompt.
Visually inspected, our method demonstrates promising performance in conditional generation, closely aligning the generated outputs with the input conditions.
For image-conditioned generation, our method yields semantically similar samples while maintaining diversity.

\cm{
For category-free text-conditioned 3D generation, we include its qualitative evaluation against state-of-the-art generic 3D generative models in Fig.~\ref{fig:t23d:oai-comparison}, along with the quantitative benchmark in Tab.~\ref{tab:t23d-clipscore}. 
As shown,
thanks to the compact 3D VAE latent space and DiT-based scalable diffusion model, our method yields better quality compared to previous diffusion-based baselines (Point-E, Shape-E, and 3DTopia) even trained with less computation resources. GaussianCube V1.1 yields sharper texture and higher aesthetic score due to the the efficient and high-resolution 3DGS rendering. Please note that GaussianCube V1.1 is trained on proprietary data and is included here for reference. More visual results of our text-conditioned \nickname{} is included in the Supp. 
}
%
%

\begin{figure*}[t!]
  \includegraphics[width=1.0\textwidth]{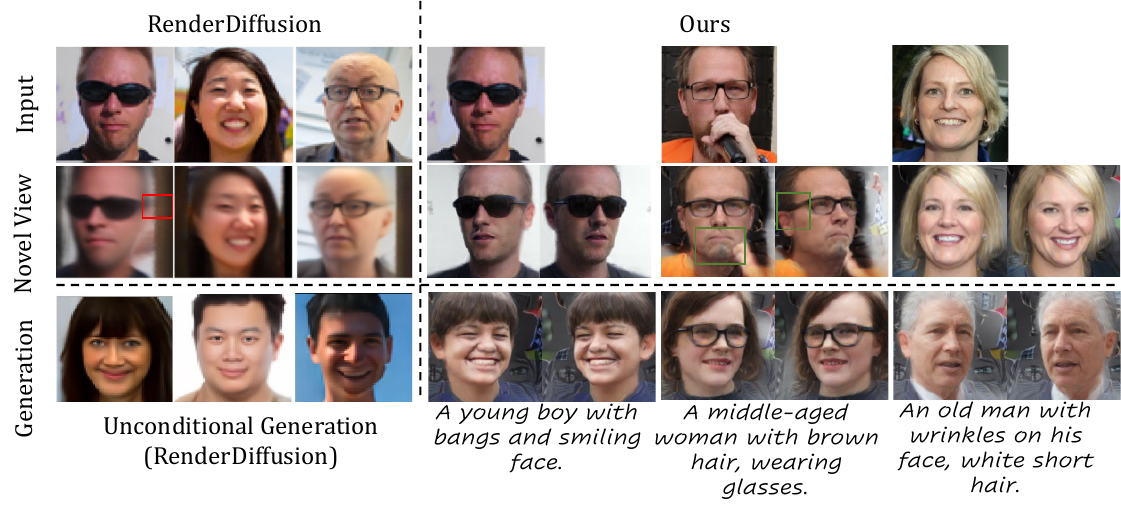} 
  \caption{\textbf{FFHQ Monocular Reconstruction (upper half) and 3D Generation (lower half).} For monocular reconstruction, we test our method with hold-out test set and visualize the input-view and novel-view. 
  Compared to baseline, our method shows consistently better performance on both reconstruction and generation.
  }
  \label{fig:ffhq}
\end{figure*}

We also compare our method against RenderDiffusion, the only 3D diffusion method that supports 3D generation over FFHQ.
As shown in the lower part of Fig.~\ref{fig:ffhq}, beyond view-consistent 3D generation, our method further supports conditional 3D generation at $128\times128$ resolution, while RenderDiffusion is limited to $64\times64$ resolution due to the expensive volume rendering integrated into diffusion training.
Quantitatively, our method achieves an FID score of $36.6$, compared to $59.3$ by RenderDiffusion.

\begin{table*}[b]
\centering
\small
\caption{
\cm{
\textbf{Quantitative evaluation of image-conditioned 3D generation.} we evaluate the quality of both 2D renderings and 3D shapes. The proposed method demonstrates strong performance across all metrics. Although multi-view images-to-3D approaches like LGM achieve better results on FID/KID metrics, they fall short on more advanced image quality assessment metrics such as CLIP-I and perform significantly worse in 3D shape quality. For multi-view to 3D methods, we also include the number of input views (V=$\#$). Note that Shape-E$^*$ is trained on proprietary internal 3D data. It is included here for reference.
}}
\resizebox{0.65\textwidth}{!}{
\begin{tabular}{lccccc}
\toprule
Method&CLIP-I$\uparrow$&FID$\downarrow$&KID(\%)$\downarrow$&COV(\%)$\uparrow$&MMD(\textperthousand)$\downarrow$   \\
\toprule
OpenLRM                        & \cellcolor{yellow}{86.37} & {38.41}  & {1.87}  & 39.33 & 29.08 \\
Splatter-Image                   & 84.10 & 48.80  & 3.65  & 37.66 & {30.69} \\
\midrule
One-2-3-45 (V=12)                     & 80.72 & 88.39  & 6.34  &  35.09 \\
CRM (V=6)                           & 85.76 & 45.53  & 1.93  &  38.83 & 28.91 \\
Lara (V=4)                          & 84.64 & 43.74  & 1.95  &  39.33 & 28.84 \\
LGM (V=4)        & \cellcolor{orange}{87.99} & \cellcolor{tablered}{19.93}  & \cellcolor{tablered}{0.55}  & {50.83} & {22.06} \\
\midrule
Shape-E$^*$                        & 77.05 & 138.53 & 11.95  & \cellcolor{tablered}{61.33} & \cellcolor{tablered}19.17 \\
\textbf{\nickname{}} & \cellcolor{tablered}{88.29} & \cellcolor{orange}{23.01}  & \cellcolor{orange}0.75  & \cellcolor{orange}55.17 & \cellcolor{orange}19.94 \\
\textbf{\nickname{}-Flexi} & {85.46} & \cellcolor{yellow}{36.28}  & \cellcolor{yellow}1.12  & \cellcolor{yellow} 53.17 & \cellcolor{yellow} 22.02 \\
\bottomrule
\end{tabular}
}
\label{tab:3d-quant-metrics}
\end{table*}

\begin{figure*}[t!]
  \includegraphics[width=1.0\textwidth]{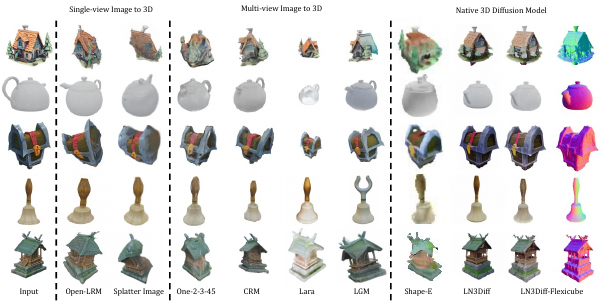} 
  \caption{\cm{
  \textbf{Qualitative Comparison of Image-to-3D}. We showcase the novel view 3D reconstruction of all methods given a single image from the unseen GSO dataset. Our proposed method achieves consistently stable performance across all cases. 
  Note that though feed-forward 3D reconstruction methods achieve sharper texture reconstruction, these methods fail to yield intact 3D predictions under challenging cases (\eg, the house in row 1). 
  In contrast, our proposed native 3D diffusion model achieves consistently better performance.
  The \nickname{}-Flexicube yields high-quality normal rendering, which cannot be easily rendered from the NeRF version.
  Zoom in for the best view.
  }
  }
  \label{fig:i23d:qualitative}
\end{figure*}

\heading{\cm{Image-to-3D Generation}} 
On the category-specific setting,
beyond the samples shown in Fig.~\ref{fig:teaser},
we include monocular reconstruction results over FFHQ datasets in the upper half of Fig.~\ref{fig:ffhq} and compare against RenderDiffusion.
As can be observed, our method demonstrates high fidelity and preserves semantic details even in self-occluded images.
The novel view generated by RenderDiffusion appears blurry and misses semantic components that are not visible in the input view, such as the leg of the eyeglass.

\cm{
On the category-free setting, our proposed framework enables 3D generation given single-view image conditions, leveraging the DiT architecture detailed in Fig.~\ref{fig:dit} (b). Following Tang~\etal~\cite{tang2024lgm}, we qualitatively benchmark our method in Fig.~\ref{fig:i23d:qualitative} over the single-view 3D reconstruction task on unseen images from the GSO dataset. Our proposed framework is robust to inputs with complicated structures (rows 1, 3, 5) and self-occlusion (row 2), yielding consistently intact 3D reconstruction. Besides, our generative-based method shows a more natural back-view reconstruction, as opposed to regression-based methods that are commonly blurry in uncertain areas.
}

\cm{
Quantitatively, we show the evaluation in Tab.~\ref{tab:3d-quant-metrics}. As can be seen, our proposed method achieves state-of-the-art performance over CLIP-I and all 3D metrics, with competitive results over conventional 2D rendering metrics FID/KID. Note that LGM leverages pre-trained MVDream~\cite{shi2023MVDream} as the first-stage generation and then maps the generated four views to pixel-aligned 3D Gaussians. This cascaded pipeline achieves better visual quality but is prone to yield distorted 3D geometry, as visualized in Fig.~\ref{fig:i23d:qualitative}.
}

\begin{figure}[h!] 
\begin{center}
  \includegraphics[width=1.0\linewidth]{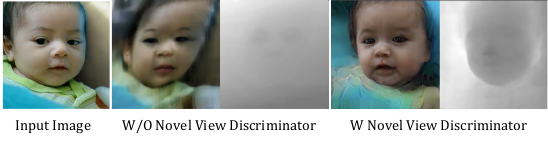} 
  \caption{Ablation of novel view discriminator. Adding novel view discriminator leads to more 3D consistent predictions.}
  \label{fig:rebuttal:cvd}
\end{center}
\end{figure}


\begin{table*}[t]
\centering
\begin{minipage}[b]{0.48\textwidth}
    \centering
    \captionof{table}{\textbf{Ablation of Reconstruction Arch Design.} We ablate the design of our auto-encoding architecture. Each component contributes to a consistent gain in the reconstruction performance, indicating an improvement in the modeling capacity.}
    \begin{tabular}{lc}
        \toprule
        Design & PSNR@100K  \\
        \midrule
        2D Conv Baseline &{17.46} \\
        \midrule
        + ViT Block &{18.92} \\
        ViT Block $\rightarrow$ DiT Block &{20.61} \\
        + Plucker Embedding &{21.29} \\
        + Cross-Plane Attention &{21.70} \\
        + Self-Plane Attention &\textbf{21.95} \\
        \bottomrule
    \end{tabular}
    \label{tab:abla:rec_arch}%
\end{minipage}
\hfill
\begin{minipage}[b]{0.49\textwidth}
   \centering 
    \captionof{table}{\textbf{Diffusion Sampling Speed and Latent Size}. We provide the sampling time per instance evaluated on $1$ V100, along with the latent size. 
    Our method achieves faster sampling speed while maintaining superior generation performance.
    }
    \begin{tabular}{lcc}
        \toprule
        Method & V100-sec &  Latent Size \\
        \midrule
        Get3D/EG3D &{$<$0.5} & 256 \\
        \midrule
        SSDNeRF &{8.1} & $128^2\times18$ \\
        RenderDiffusion & {15.8} & - \\
        DiffRF & {18.7} & $32^3 \times 4$ \\
        \nickname{}$_\text{uncond}$ &\textbf{5.7} &  $32^2\times12$ \\
        \nickname{}$_\text{cfg}$ & \textbf{7.5} & $32^2\times12$ \\
        \bottomrule
    \end{tabular}
    \label{tab:exp:sampling_speed_and_latent_size}%
\end{minipage}
\end{table*}

\subsection{Ablation Study and Analysis}
\noindent\textbf{Reconstruction Arch Design}.
In Tab.~\ref{tab:abla:rec_arch}, we benchmark each component of our auto-encoding architecture over a subset of Objaverse with $7K$ instances and record the PSNR at $100K$ iterations.
Each component introduces consistent improvements with negligible parameter increases.

\noindent\textbf{Novel View Discriminator for Monocular Dataset.}
Novel view discriminator is crucial for monocular datasets like FFHQ. 
As shown in Fig.~\ref{fig:rebuttal:cvd}, without it, the VAE model fails to yield a plausible novel view.

\noindent\textbf{Diffusion Sampling Speed and Latent Size.}
We report the sampling speed and latent space size comparison in Tab.~\ref{tab:exp:sampling_speed_and_latent_size}.
By performing on the compact latent space, our method achieves the fastest sampling while keeping the best generation performance.
Though RenderDiffusion follows a latent-free design, its intermediate 3D neural field has a shape of $256^2\times96$ and hinders efficient diffusion training.

\begin{figure}[h!] 
\begin{center}
  \includegraphics[width=1.0\linewidth]{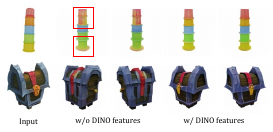} 
  \caption{Ablation of DINO features in image-conditioned 3D generation. Using CLIP features only leads to unfaithful 3D generation, as highlighted in the red box. Introducing auxiliary DINO features improves both fidelity and faithfulness.}
  \label{fig:abla:dino-clip-feat}
\end{center}
\end{figure}

\heading{DINO Features for Image-conditioned 3D Generation}
As shown in Fig.~\ref{fig:abla:dino-clip-feat}, we ablate the DINO features in image-conditioned 3D generation. As can be seen, using CLIP features only as the condition leads to unfaithful 3D generation, where the generated asset does not correctly reflect the input condition. Introducing low-level DINO features effectively resolves this issue.
\cmfinal{
Note that we adopt the prepend-based image conditioning mechanism in Fig.\ref{fig:dit}, motivated by its proven effectiveness in SD-3\cite{esser2024sd3}. While our follow-up work GaussianAnything~\cite{lan2024ga} shows that cross-attention–based conditioning offers comparable performance with lower VRAM cost, we prioritize the prepend-based design here for its simplicity and demonstrated reliability.}


\section{Conclusion and Discussions}
In this work, we introduce a new paradigm for 3D generative modeling by training a diffusion model over a compact, 3D-aware latent space.. 
A dedicated variational autoencoder encodes (multi-view) image(s) into a low-dimensional structured latent space, where conditional diffusion learning can be efficiently performed.
We achieve state-of-the-art performance over ShapeNet and demonstrate our method over generic category-free Objaverse 3D datasets.
Our work can facilitate numerous downstream applications in 3D vision and graphics tasks.

\vspace{0.1cm}
\heading{Limitations and Future Work}
\cm{
Our method still has several unresolved limitations.
\textbf{From the VAE perspective}, we observe that volume rendering remains memory-intensive. In addition, the visual quality of the fine-tuned Flexicube variant is inferior to that of the NeRF-based version. Extending our decoder to more efficient 3D representations, such as 2DGS~\cite{Huang2DGS2024}, may enable faster rendering while preserving high-quality 3D surfaces.
%
Moreover, our triplane-structured 3D VAE latent space may not be the optimal choice for learning 3D diffusion. More explicit 3D representations, such as point clouds or sparse voxels~\cite{spconv2022}, could also serve as effective latent proxies for 3D VAE training. Besides, directly incorporating 3D supervision, such as SDF or occupancy signals, may further improve geometric reconstruction performance.
%
\cmfinal{
In addition, training our proposed model on single-view datasets (\eg, FFHQ) leads to unnatural background patterns, as illustrated in Fig.~\ref{fig:ffhq}. We hypothesize that these artifacts arise from the novel view discriminator: while it enforces plausible novel-view rendering, the background region often develops artifacts that function as \emph{adversarial shortcuts} to increase the discriminator score. A potential solution to mitigate this artifact is to reduce the GAN loss weights of the novel view discrimination training, but at the cost of 3D plausibility~\ref{fig:rebuttal:cvd}. Therefore, developing more effective ways for training 3D diffusion models over a monocular view dataset is a valuable avenue for future research.
}
\textbf{On the diffusion formulation}, \nickname{} learns the joint distribution of geometry and texture. This is likely to yield suboptimal performance. A geometry-texture disentangled framework~\cite{zhang2024clay} enables more flexible generation and better texture quality by leveraging the powerful priors of 2D diffusion model~\cite{rombach2022LDM}.
\textbf{Regarding data},~\nickname{} is trained on the artist-created data only. Adding more real-world data such as MVImageNet~\cite{yu2023mvimgnet} shall further improve the generality. Besides, Objaverse-XL~\cite{objaverseXL} and 3D-Future~\cite{fu20213d} may enhance model diversity.
\textbf{Application side}, Current 3D generation research remains largely focused on base model design. Future work could explore incorporating 3D-aware control signals~\cite{zhang2023adding}, toonification~\cite{zhang2023deformtoon3d,yang2022vtoonify} and training-free editing~\cite{mou2023diffeditor}.
Overall, our method represents a step toward a native 3D diffusion model and offers inspiration for future research in this direction.
}

\vspace{0.1cm}
\heading{Acknowledgement}
This study is supported under the RIE2020 Industry Alignment Fund Industry Collaboration Projects (IAF-ICP) Funding Initiative, as well as cash and in-kind contributions from the industry partner(s). It is also supported by Singapore MOE AcRF Tier 2 (MOE-T2EP20221-0011).

\appendix

In this supplementary material, we provide additional details regarding the implementations and additional results. We also discuss the limitations of our model.
\vspace{2mm}

\heading{Broader Social Impact} 
In this paper, we introduce a new latent 3D diffusion model designed to produce high-quality textures and geometry using a single model. 
As a result, our approach has the potential to be applied to generating DeepFakes or deceptive 3D assets, facilitating the creation of falsified images or videos. 
This raises concerns as individuals could exploit such technology with malicious intent, aiming to spread misinformation or tarnish reputations.


\section{Implementation details}
\subsection{Training details} 

\heading{Diffusion}
We mainly adopt the diffusion training pipeline implementation from ADM~\cite{dhariwal2021diffusion}, continuous noise schedule from LSGM~\cite{vahdat2021score} with the spatial transformer attention implementation from LDM~\cite{rombach2022LDM}.
For ShapeNet and FFHQ dataset, we adopt U-Net~\cite{ronneberger_u-net_2015} architecture and list the hyperparameters in Tab.~\ref{tab:diffusion-params}.
For Objaverse dataset, we adopt DiT-L~\cite{Peebles2022DiT} architecture with cross attention design, as proposed in PixArt~\cite{chen2023pixartalpha}. 
The diffusion transformer is built with $24$ layers with $16$ heads and $1024$ hidden dimension, which result in $458$M parameters.

\begin{table}%
\caption{Hyperparameters and architecture of diffusion model $\model$. 
}
\label{tab:diffusion-params}
\begin{minipage}{\columnwidth}
\begin{center}
\begin{tabular}{ll}
\toprule
Diffusion Model Details \\
\midrule
Learning Rate &  $2e-5$ \\
Batch Size &  $96$ \\
Optimizer &  AdamW \\
Iterations &  500K \\
\midrule
U-Net base channels &  320 \\
U-Net channel multiplier &  1, 1, 2, 2, 4, 4 \\
U-Net res block &  2 \\
U-Net attention resolutions &  4,2,1 \\
U-Net Use Spatial Transformer &  True \\
U-Net Learn Sigma &  False \\
U-Net Spatial Context Dim &  768 \\
U-Net attention head channels &  64 \\
U-Net pred type &  $v$ \\
U-Net norm layer type &  GroupNorm \\
\midrule
Noise Schedules & Linear \\
CFG Dropout prob & $15\%$ \\
CLIP Latent Scaling Factor & $18.4$ \\
\bottomrule
\end{tabular}
\end{center}
\bigskip\centering
\end{minipage}
\vspace{-4mm}
\end{table}%
\heading{VAE Architecture} 
For the convolutional encoder $\ldmencoder$, we adopt a lighter version of LDM~\cite{rombach2022LDM} encoder with channel $64$ and $1$ residual blocks for efficiency.
When training on Objaverse with $V=6$, we incorporate 3D-aware attention~\cite{shi2023MVDream} in the middle layer of the convolutional encoder.
For convolutional upsampler $\mathcal{D}_U$, we further half the channel to $32$. All other hyper-parameters remain at their default settings.
Regarding the transformer decoder $\mathcal{D}_T$, we employ the DiT-L/2 architecture, and overall saved VAE model takes around $1.5$ GiB storage.
The input dimension of $z$ to the MLP in each DiT block is $h \times w \times c$ for self-plane attention, and $h \times w \times 3 \times c$ in cross-plane attention.
When ablating the 3D-aware attention in Tab.3, we adopt channel-wise concatenated latent $h \times w \times (3c)$ for model input, as in SSDNeRF.
Note that we trade off a smaller model with faster training speed due to the overall compute limit, and a heavier model would certainly empower better performance~\cite{Peebles2022DiT,xu2023dmv3d}.
We ignore the plucker camera condition for the ShapeNet and FFHQ dataset, over which we find raw RGB input already yields good enough performance.

\subsection{Data and Baseline Comparison}
\heading{Training data} For ShapeNet, following GET3D~\cite{gao2022get3d}, we use the blender to render the multi-view images from 50 viewpoints for all ShapeNet datasets with foreground mask. Those camera points sample from the upper sphere of a ball with a 1.2 radius.
For Objaverse, we use a high-quality subset from the pre-processed rendering from G-buffer Objaverse~\cite{qiu2023richdreamer} for experiments. 
Since G-buffer Objaverse splits the subset into $10$ general categories, we use all the 3D instances except from ``Poor-quality'': Human-Shape, Animals, Daily-Used, Furniture, Buildings$\&$Outdoor, Transportations, Plants, Food and Electronics.
The ground truth camera pose, rendered multi-view images and depth maps are used for stage-1 VAE training.

\heading{Evaluation} The 2D metrics are calculated between 50k generated images and all available real images. Furthermore, for comparison of the geometrical quality, we sample 4096 points from the surface of 5000 objects and apply the Coverage Score (COV) and Minimum Matching Distance (MMD) using Chamfer Distance (CD) as follows:
\begin{equation}
\begin{aligned}
&CD(X,Y)=\sum\limits_{x\in X} \mathop{min}\limits_{y\in Y}||x-y||^2_2+\sum\limits_{y\in Y} \mathop{min}\limits_{x\in X}||x-y||^2_2,\\
&COV(S_g,S_r)=\dfrac{|\{\mathrm{\mathop{arg~min}}_{Y \in S_r}CD(X,Y)|X\in S_g\}|}{|S_r|}\\
&MMD(S_g,S_r)=\dfrac{1}{|S_r|}\sum\limits_{Y\in S_r}\mathop{min}\limits_{X\in S_g}CD(X,Y)
\end{aligned}
~ ,
\end{equation}
where $X \in S_g$ and $Y \in S_r$ represent the generated shape and reference shape.

Note that we use 5k generated objects $S_g$ and all training shapes $S_r$ to calculate COV and MMD. For fairness, we normalize all point clouds by centering in the original and recalling the extent to [-1,1]. Coverage Score aims to evaluate the diversity of the generated samples, and MMD is used for measuring the quality of the generated samples. 2D metrics are evaluated at a resolution of 128 $\times$ 128. Since the GT data contains intern structures, we only sample the points from the outer surface of the object for results of all methods and ground truth.
\begin{figure*}[t!]
  \centering
  \includegraphics[width=0.9\textwidth]{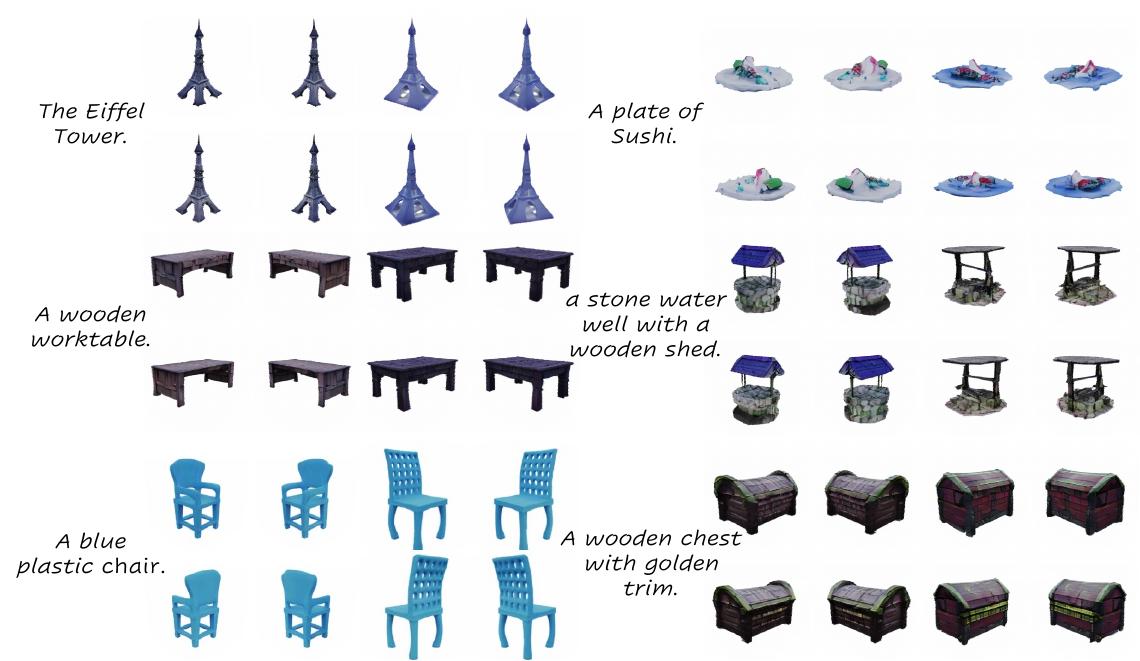} 
  \caption{\textbf{Objaverse Conditional Generation Given Text Prompt.} We show two samples for each prompt. All results are sampled from DiT architectures. Zoom in for the best view.}
  \label{fig:objaverse:cond}
\end{figure*}

For FID/KID evaluation, since different methods have their unique evaluation settings, we standardize this process by re-rendering each baseline's samples using a fixed upper-sphere ellipsoid camera pose trajectory of size $20$.
With $2.5K$ sampled 3D instances for each method, we recalculate FID@50K/KID@50K, ensuring a fair comparison across all methods.
\cm{For text-conditioned 3D generation on category-free datasets, we leverage the text prompt provided in GPTEval3D~\cite{wu2023gpteval3d} for evaluation. The CLIP similarity score is calculated following previous work~\cite{hong20243dtopia}. Note that GaussianCube V1.1 is trained on proprietary data, while our method is trained on public dataset. We failed to run GaussianCube V1.0 since the publicly released checkpoint is broken.}

\heading{Details about Baselines}
We reproduce EG3D, GET3D, and SSDNeRF on our ShapeNet rendering using their officially released codebases.
In the case of RenderDiffusion, we use the code and pre-trained model
shared by the author for ShapeNet experiments. 
Regarding FFHQ dataset, 
due to the unavailability of the corresponding inference configuration and checkpoint from the authors,
we incorporate their unconditional generation and monocular reconstruction results as reported in their paper.
For DiffRF, 
given the absence of the public code,
we reproduce their method with Plenoxel~\cite{yu_and_fridovichkeil2021plenoxels} and ADM~\cite{dhariwal2021diffusion}.

\section{More Results}
\subsection{More Qualitative 3D Generation Results}
\begin{figure*}[t!]
  \includegraphics[width=1.0\textwidth]{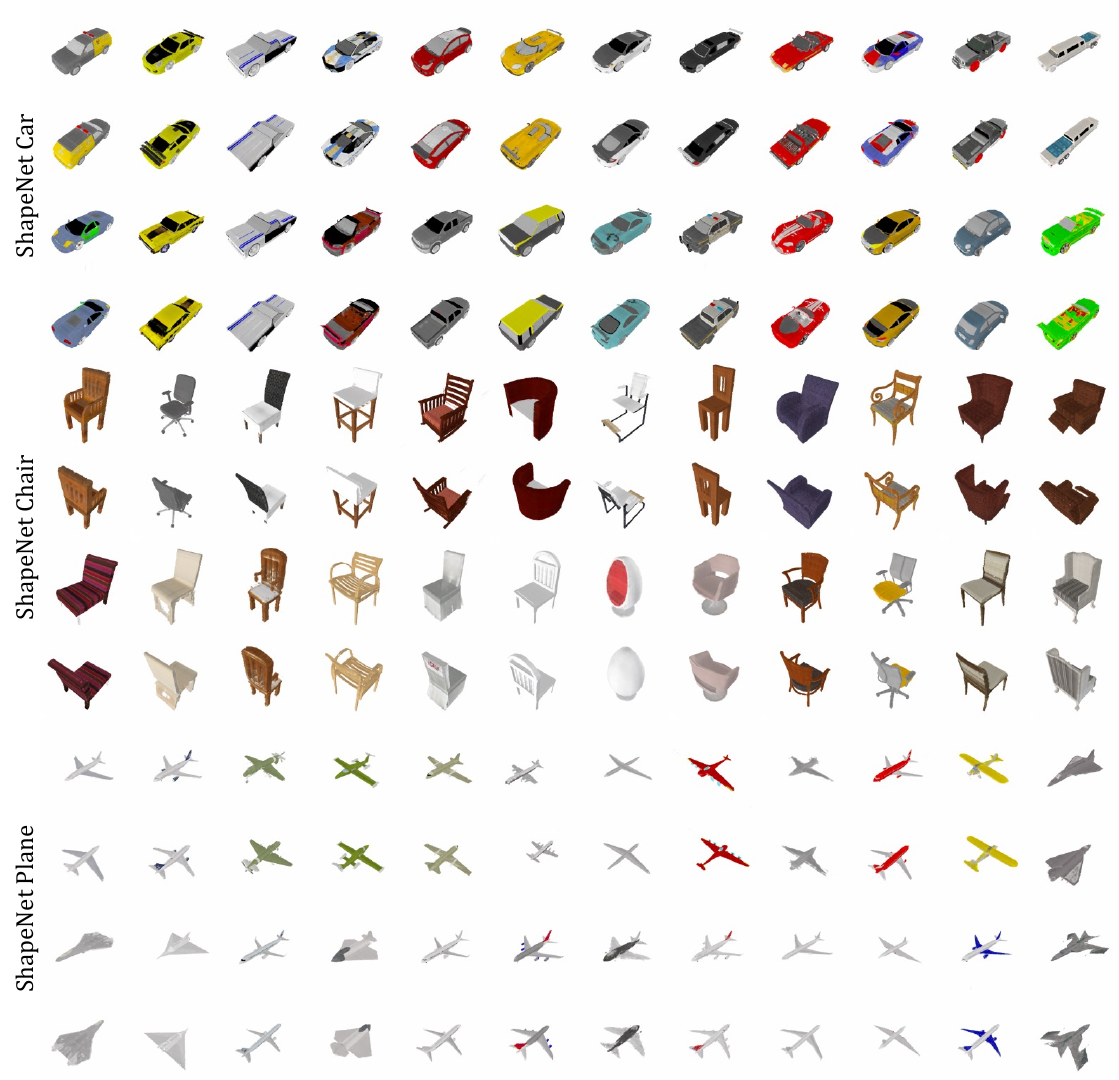} 
  \caption{\textbf{Unconditional 3D Generation by \nickname{} (Uncurated).} We showcase uncurated samples generated by \nickname{} on ShapeNet three categories. We visualize two views for each sample. Better zoom in.}
  \label{fig:shapenet:uncurated}
\end{figure*}
\begin{figure*}[t!]
  \includegraphics[width=1.0\textwidth]{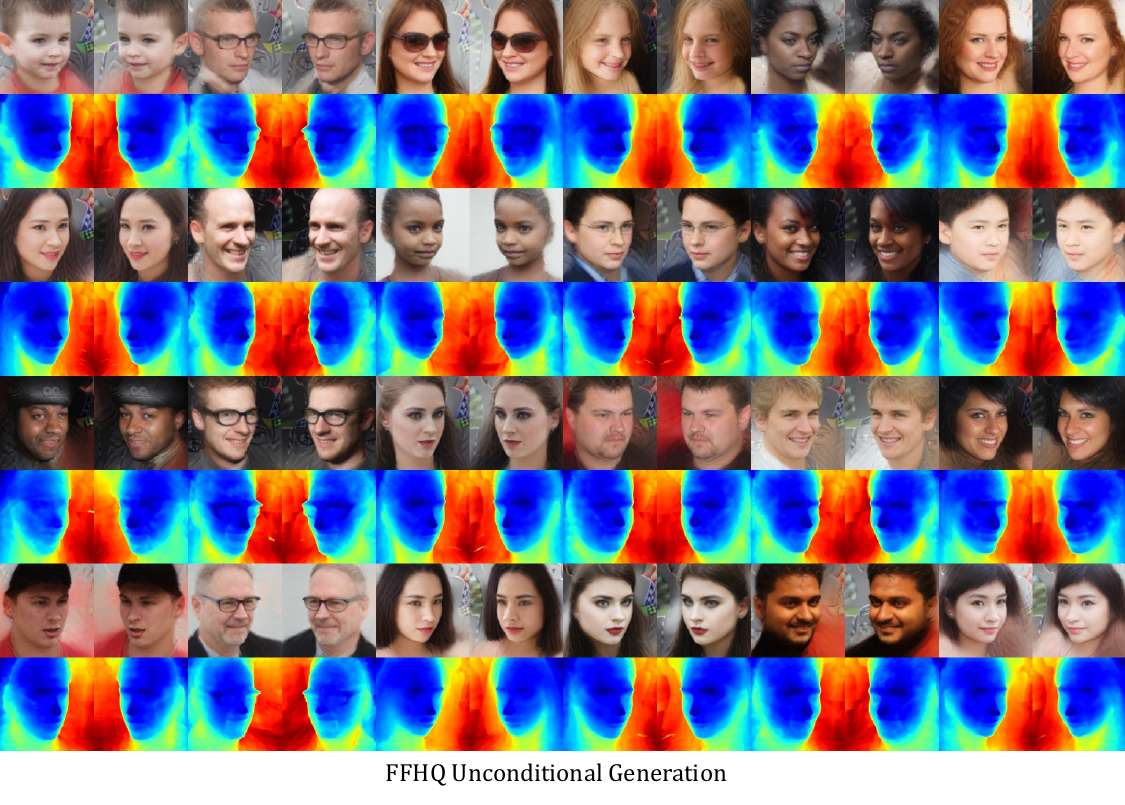} 
  \caption{\textbf{Unconditional 3D Generation by \nickname{} (Uncurated).} We showcase uncurated samples generated by \nickname{} on FFHQ. We visualize two views for each sample along with the extracted depth. Better zoom in.}
  \label{fig:ffhq:uncurated}
\end{figure*}
We include more uncurated samples generated by our method on ShapeNet in Fig.~\ref{fig:shapenet:uncurated}, on FFHQ in Fig.~\ref{fig:ffhq:uncurated}, and category-free text-conditioned 3D generation results in Fig.~\ref{fig:objaverse:cond}

%

\subsection{More Monocular 3D Reconstruction Results}
We further benchmark the generalization ability of our stage-1 monocular 3D reconstruction VAE.
\begin{table}[t!]
\centering
\caption{
Quantitative results on ShapeNet-SRN~\cite{shapenet2015, sitzmann2019srns} chairs evaluate on $128\times128$. 
Legend: * -- requires test time optimization.
Note that our stage-1 VAE shares the same setting only with Pix2NeRF~\cite{yu2021pixelnerf}, which also has an explicit latent space for generative learning.
Other baselines are included for reference.
} 
\begin{tabular}{lcc}
\toprule
Method & PSNR $\uparrow$ & SSIM $\uparrow$\\
\midrule
GRF~\cite{grf2020} & 21.25 & 0.86 \\
TCO~\cite{tatarchenko2016tco} & 21.27 & 0.88 \\
dGQN~\cite{Eslami2018NeuralSR} & 21.59 & 0.87 \\
ENR~\cite{dupont2020enr} & 22.83 & - \\
SRN*~\cite{sitzmann2019srns} & 22.89 & 0.89 \\
CodeNeRF*~\cite{jang2021codenerf} & 22.39 & 0.87 \\
PixelNeRF~\cite{yu2021pixelnerf} & \textbf{23.72} & \textbf{0.91} \\
\midrule
Pix2NeRF~\cite{cai2022pix2nerf} conditional & 18.14 & 0.84 \\
Ours & 20.91 & 0.89 \\
\bottomrule
\end{tabular}
%
%
\label{tab:srn_chairs}
\end{table}%
For ShapeNet, we include the quantitative evaluation in Tab.~\ref{tab:srn_chairs}.
Our method achieves a comparable performance with monocular 3D reconstruction baselines.
Note that strictly saying, our stage-1 VAE shares a similar setting with Pix2NeRF~\cite{cai2022pix2nerf}, whose encoder also has a latent space for generative modeling.
Other reconstruction-specific methods like PixelNeRF~\cite{yu2021pixelnerf} do not have these requirements and can leverage some designs like pixel-aligned features and long-skip connections to further boost the reconstruction performance.
We include their performance mainly for reference and leave training the stage-1 VAE model with performance comparable with those state-of-the-art 3D reconstruction models for future work.

\begin{figure}[t]
\begin{minipage}{0.4\textwidth}
   \hspace{0.05\textwidth} 
  \includegraphics[width=1.0\textwidth]{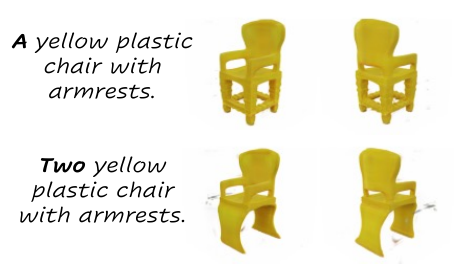} 
\end{minipage}
\hspace{0.05\textwidth}
\begin{minipage}{0.4\textwidth}
   \hspace{0.05\textwidth} 
  \caption{\textbf{Limitation analysis.} We showcase the deficiency to generate composed 3D scenes by \nickname{}. As shown here, the prompt $\textbf{Two}$ chair yields similar results with $\textbf{A}$ chair.}
  \label{fig:abla:compositionality}
\end{minipage}
\end{figure}

\begin{figure*}[h!]
  \includegraphics[width=0.9\textwidth]{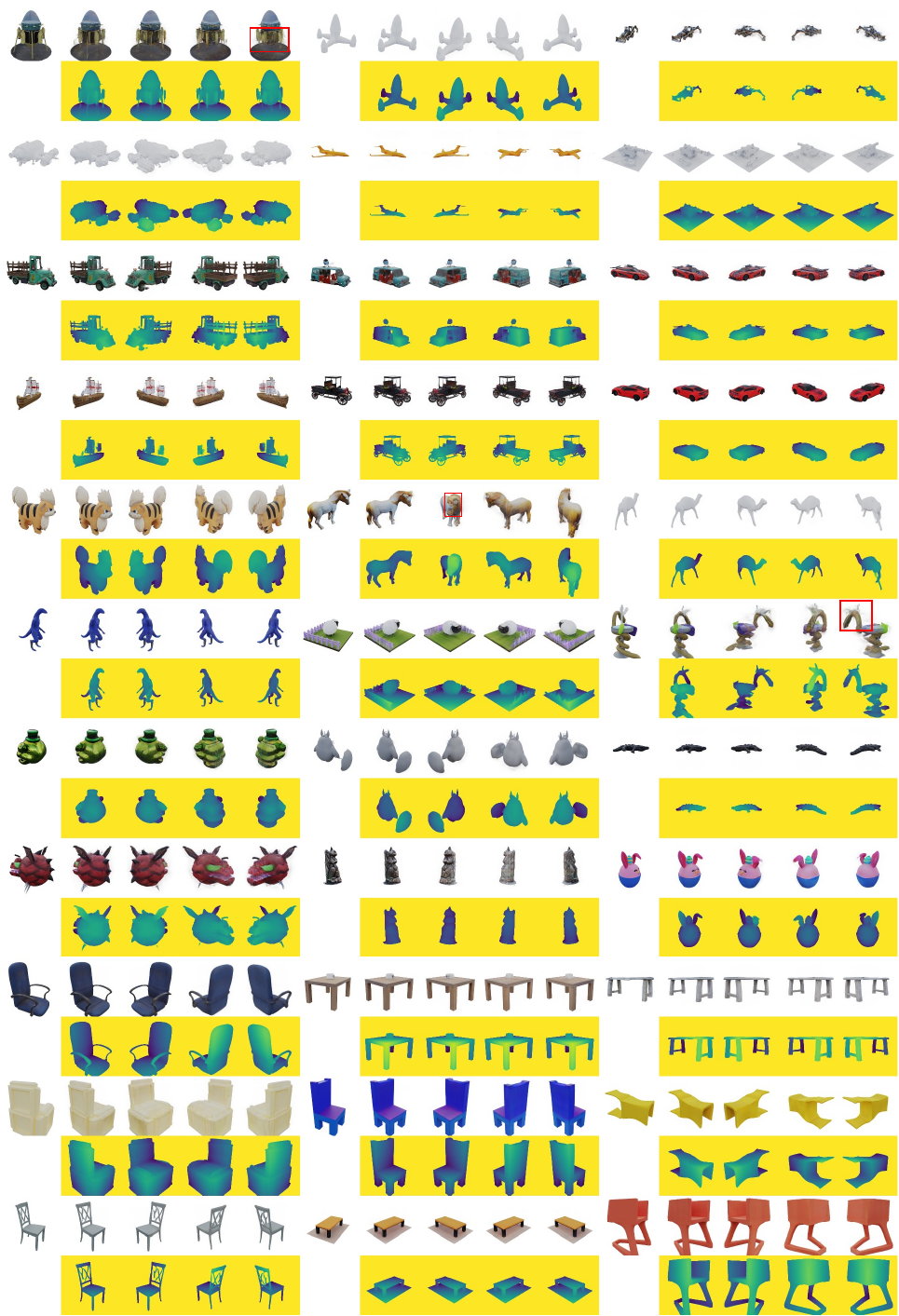} 
  \caption{\textbf{Monocular 3D Reconstruction by \nickname{} stage-1 VAE on Objaverse (Uncurated).} We showcase uncurated samples monocular-reconstructed by \nickname{} on Objaverse. From left to right, we visualize the input image, four reconstructed novel views with the corresponding depth maps. 
  Artifacts are labeled in 
  {\textcolor{red}{Red}}. 
  Better zoom in.}
  \label{fig:objaverse:uncurated}
\end{figure*}
Besides, we visualize \nickname{}'s stage-1 monocular VAE reconstruction performance over our Objaverse split in Fig.~\ref{fig:objaverse:uncurated}.
As can be seen, though only one view is provided as the input, our monocular VAE reconstruction can yield high-quality and view-consistent 3D reconstruction with a detailed depth map.
Quantitatively, the novel-view reconstruction performance over our whole Objaverse dataset achieves an average PSNR of $26.14$.
This demonstrates that our latent space can be treated as a compact proxy for efficient 3D diffusion training.

\section{Limitation and Failure Cases}
We have included a brief discussion of limitations in the main submission. 
Here we include more details along with the visual failure cases for a more in-depth analysis of \nickname{}'s limitations and future improvement directions. 

\subsection{VAE Limitations}
We have demonstrated that using a monocular image as encoder input can achieve high-quality 3D reconstruction.
However, we noticed that for some challenging cases with diverse color and geometry details, the monocular encoder leads to blurry artifacts.
As labeled in Fig.~\ref{fig:objaverse:uncurated}, our method with monocular input may yield floating artifacts over unseen viewpoints.
We hypothesize that these artifacts are largely due to the ambiguity of monocular input and the use of regression loss (L2/LPIPS) during training.
These observations demonstrate that switching to a multi-view encoder is necessary for better performance.

Besides, since our VAE requires plucker camera condition as input, the pre-trained VAE method cannot be directly applied to the unposed dataset.
However, we believe this is not a research issue at the current time, considering the current methods still perform lower than expected on existing high-quality posed 3D datasets like Objaverse.

\subsection{3D Diffusion Limitations}
As one of the earliest 3D diffusion models that works on Objaverse, our method still suffers from several limitations that require investigation in the future.
\textbf{(1)} The support of image-to-3D on Objaverse. Currently, we leverage CLIP$_\text{text}$ encoder with the $77$ tokens as the conditional input. However, unlike 2D AIGC with T2I models~\cite{rombach2022LDM}, 3D content creation can be greatly simplified by providing easy-to-get 2D images. 
An intuitive implementation is by using our ShapeNet 3D diffusion setting, which provides the final normalized CLIP text embeddings as the diffusion condition.
However, as shown in the lower half of Fig.~4 in the main submission, the CLIP encoder is better at extracting high-level semantics rather than low-level visual details.
Therefore, incorporating more accurate image-conditioned 3D diffusion design like ControlNet~\cite{zhang2023adding} to enable monocular 3D reconstruction and control is worth exploring in the future.
\textbf{(2)} Compositionality. Currently, our method is trained on object-centric dataset with simple captions, so the current model does not support composed 3D generation.
For example, the prompt "Two yellow plastic chair with armchests" will still yield one chair, as visualized in Fig.~\ref{fig:abla:compositionality}.
\textbf{(3)} UV map. To better integrate the learning-based method into the gaming and movie industry, a high-quality UV texture map is required. 
A potential solution is to disentangle the learned geometry and texture space and build the connection with UV space through dense correspondences~\cite{lan2022ddf_ijcv}.



\clearpage

\subsection{Biographies}
\begin{IEEEbiography}[{\includegraphics
[width=1in,height=1.25in,clip,
keepaspectratio]{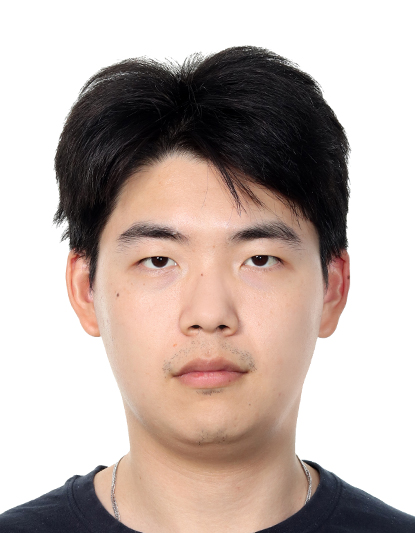}}]{Yushi Lan}
is a postdoctoral research fellow at the Visual Geometry Group, University of Oxford. He received his Ph.D. degree from the College of Computing and Data Science, Nanyang Technological University, in 2025, under the supervision of Prof. Chen Change Loy. He received his B.Eng. degree in Software Engineering from Beijing University of Posts and Telecommunications, China, in 2020. His research focuses on the intersection of computer vision, graphics, and machine learning, with particular interests in 3D generative models and 3D representation learning. He has served as a reviewer for top-tier conferences and journals, including CVPR, ICCV, ECCV, ICLR, NeurIPS, and TPAMI.
\end{IEEEbiography}

\begin{IEEEbiography}[{\includegraphics
[width=1in,height=1.25in,clip,
keepaspectratio]{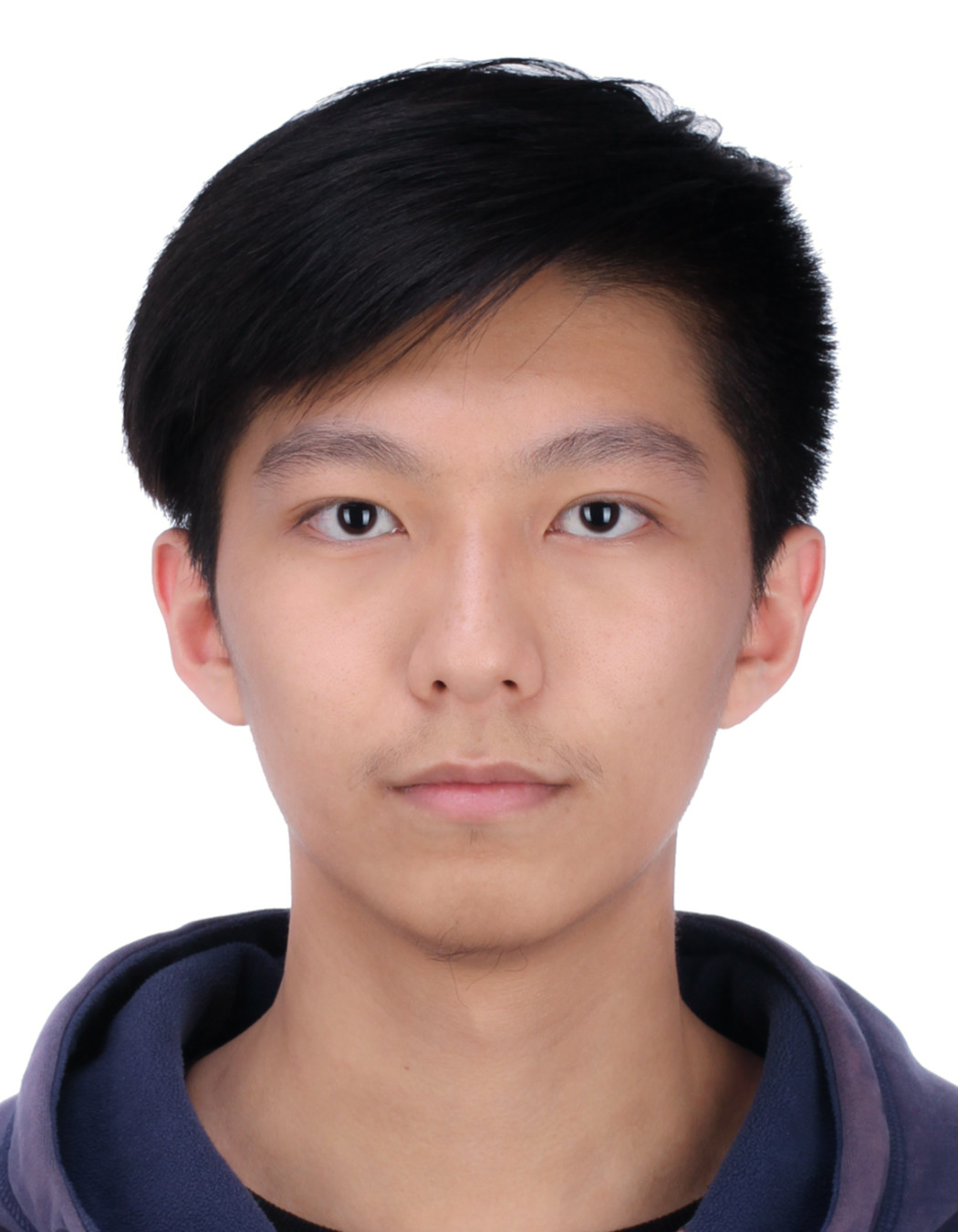}}]{Fangzhou Hong}
   is a Research Fellow in the College of Computing and Data Science at Nanyang Technological University, advised by Prof. Ziwei Liu. He received Ph.D. degree in the College of Computing and Data Science at Nanyang Technological University in 2025. He received the B.Eng. degree in Software Engineering from Tsinghua University, China, in 2020. His research interests lie on computer vision and deep learning. Particularly, he is interested in 3D representation learning. He has published several papers and served as a reviewer for top conferences and journals, \eg, CVPR, ICCV, ECCV, ICLR, NeurIPS, TPAMI.
\end{IEEEbiography}

\begin{IEEEbiography}[{\includegraphics
[width=1in,height=1.25in,clip,
keepaspectratio]{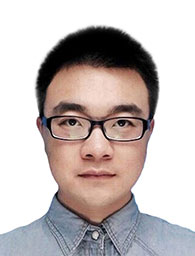}}]{Shangchen Zhou}
is currently a Research Assistant Professor at MMLab@NTU, Nanyang Technological University, Singapore. He received his Ph.D. (2024) in Computer Science from the same institution. He was selected as an outstanding reviewer in NeurIPS 2020. He won first place in three image restoration and enhancement challenges in NTIRE 2021. His works received notable recognition including the WAIC Youth Outstanding Paper Award Honorable Mention in 2023, the Snap Fellowship Honorable Mention in 2022, and the Best Paper Award at ICIMCS, ACM in 2016. Additionally, He co-organized the ``MIPI workshop'' series in conjunction with ECCV 2022, CVPR 2023, and CVPR 2024. His research interests include image/video restoration and enhancement, generation and editing, etc.
\end{IEEEbiography}

\begin{IEEEbiography}[{\includegraphics
[width=1in,height=1.25in,clip,
keepaspectratio]{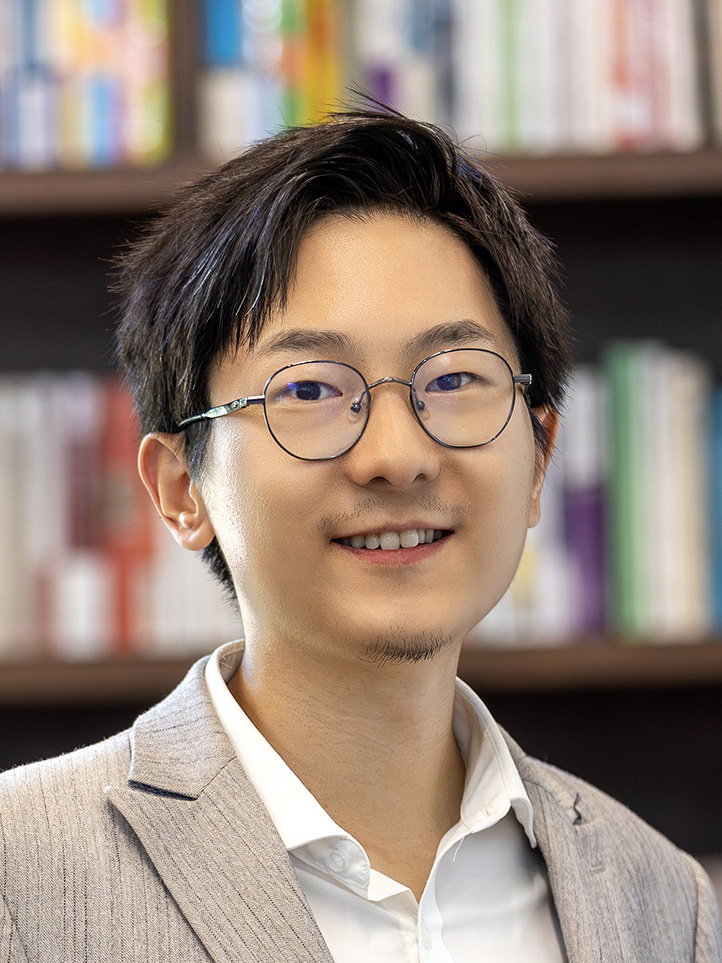}}]{Shuai Yang}(S'19-M'20) 
received the B.S. and Ph.D. degrees (Hons.) in computer science from Peking University, Beijing, China, in 2015 and 2020, respectively. He is currently an assistant professor with the Wangxuan Institute of Computer Technology, Peking University. His current research interests include image stylization, image translation and image editing. He was a Research Assistant Professor with the S-Lab, Nanyang Technological University, Singapore, from Mar. 2023 to Feb. 2024. He was a postdoctoral research fellow at Nanyang Technological University, from Oct. 2020 to Feb. 2023. He was a Visiting Scholar with the Texas A\&M University, from Sep. 2018 to Sep. 2019. He was a Visiting Student with the National Institute of Informatics, Japan, from Mar. 2017 to Aug. 2017. He received the IEEE ICME 2020 Best Paper Awards and IEEE MMSP 2015 Top10\% Paper Awards.  He  serves/served as an Area Chair of NeurIPS, ACM MM, BMVC, CVPR and ICLR.

\end{IEEEbiography}

\begin{IEEEbiography}[{\includegraphics
[width=1in,height=1.25in,clip,
keepaspectratio]{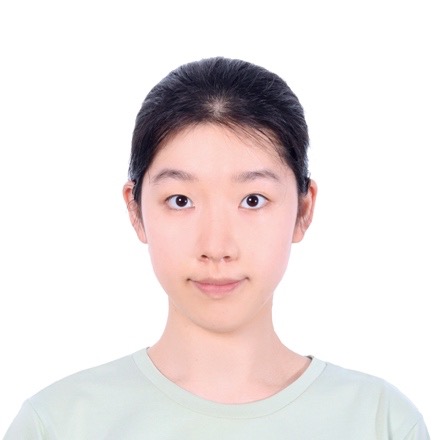}}]{Xuyi Meng} is a Ph.D. student in Computer Science at Cornell University, advised by Prof. Wei-Chiu Ma. She received the M.S. degree in Computer and Information Science from the University of Pennsylvania in 2025 and the B.E. degree in Computer Science from Nanyang Technological University (NTU), Singapore, in 2023.
Her research interests include 3D and 4D computer vision, generative modeling, and neural representations, with a focus on efficient and high-quality modeling of real-world scenes and dynamics.

\end{IEEEbiography}

\begin{IEEEbiography}[{\includegraphics
[width=1in,height=1.25in,clip,
keepaspectratio]{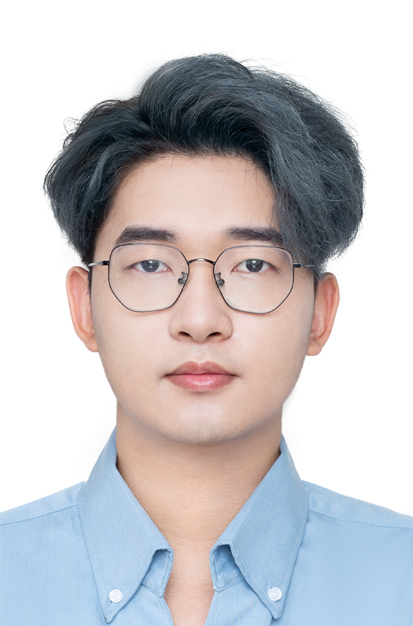}}]
{Yongwei Chen} is a Ph.D. student at the College of Computing and Data Science, Nanyang Technological University, Singapore. His research focuses on 3D vision, with an emphasis on generative 3D modeling, neural rendering, and 3D understanding. His work has been published in top-tier conferences such as CVPR, ICCV, ECCV, and NeurIPS, contributing to advancements in 3D computer vision and its applications.
\end{IEEEbiography}

\begin{IEEEbiography}[{\includegraphics
[width=1in,height=1.25in,clip,
keepaspectratio]{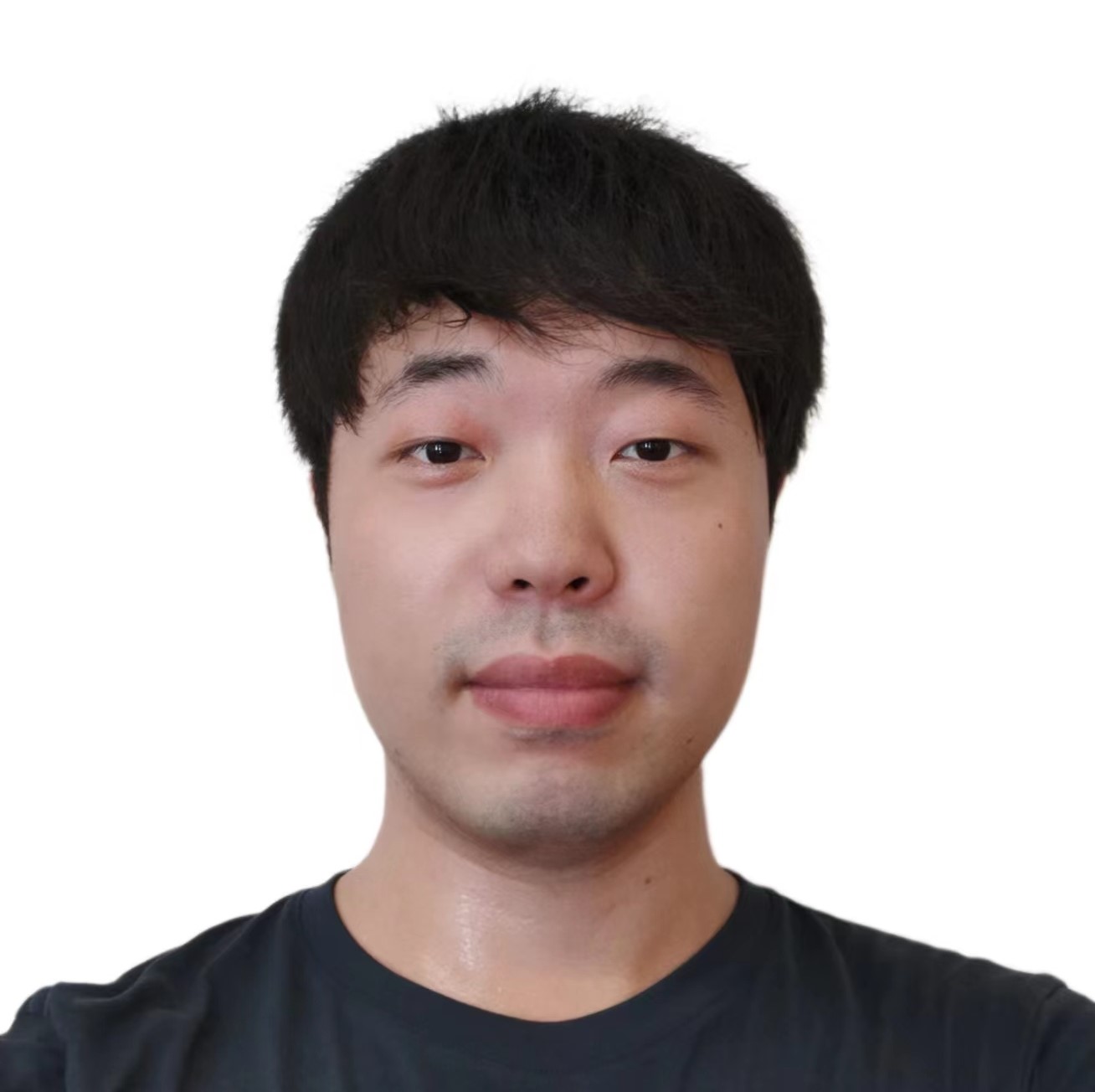}}]
{Zhaoyang Lyu}
is a research scientist with the Shanghai AI Laboratory, working in a research group on Embodied AI. 
He received his Ph.D. (2018-2022) from Multimedia Laboratory (MMLab) at CUHK, advised by Prof. Dahua Lin. He obtained his Bachelor's Degree (2014-2018) at Xi'an Jiaotong University. His current research focuses on generative world modeling for Embodied AI, including 3D interactive object and scene generation. 
\end{IEEEbiography}

\begin{IEEEbiography}[{\includegraphics
[width=1in,height=1.25in,clip,
keepaspectratio]{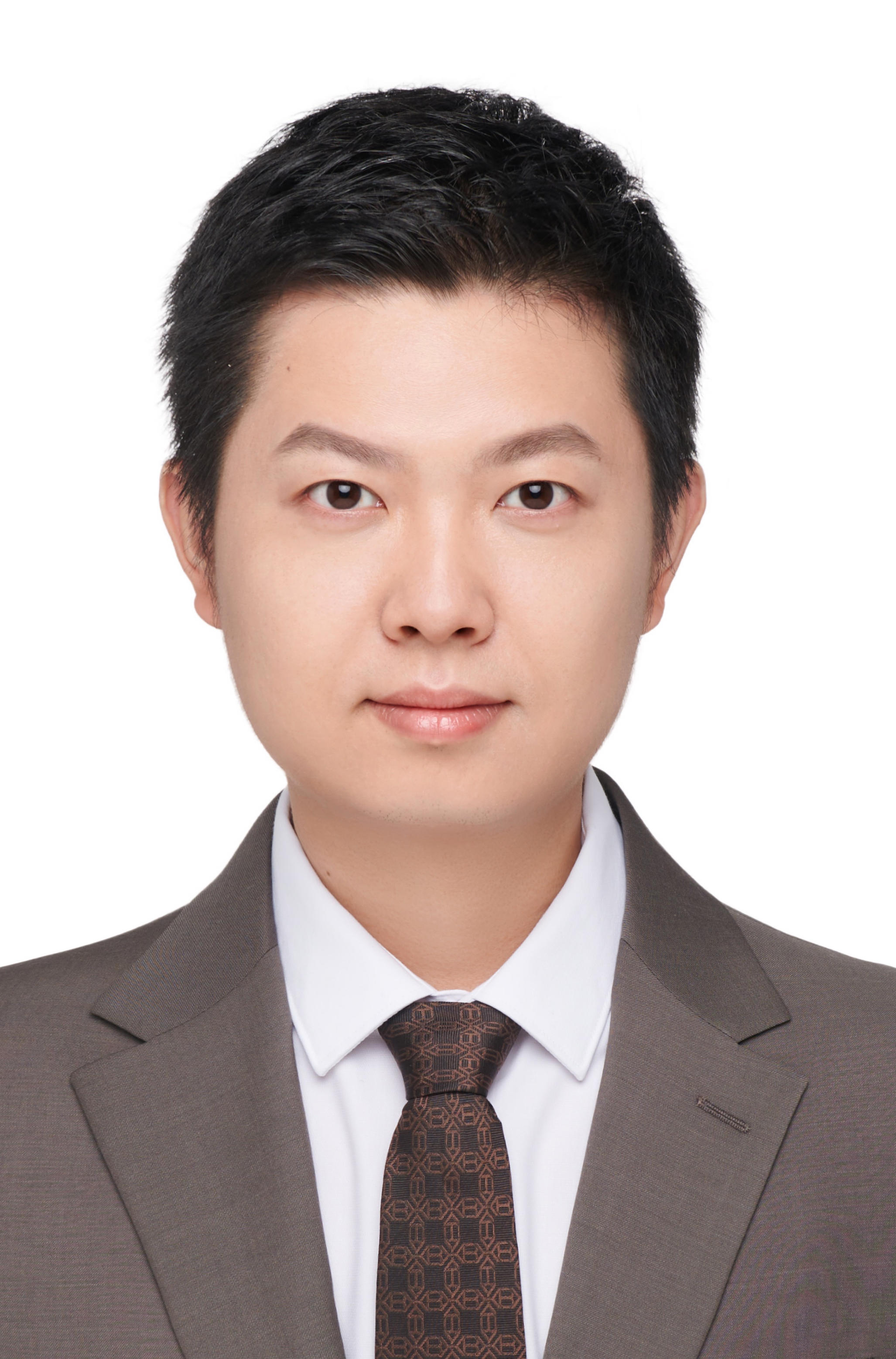}}]
{Bo Dai}
is an assistant professor in the Musketeers Foundation Institute of Data Science, The University of Hong Kong. He obtained his PhD degree from The Chinese University of Hong Kong. He was a research scientist with the Shanghai Artificial Intelligence Laboratory, and was a research assistant professor with S-Lab for advanced intelligence, Nanyang Technological University. He has authored or coauthored more than 80 papers in top-tier conferences and journals, with over 14000 google scholar citations. His research interests include Generative AI and its interdisciplinary applications in areas covering Embodied AI, Scientific Discovery, Metaverse and Creativity. He is an area chair of ICLR, CVPR, NeurIPS and AAAI.
\end{IEEEbiography}

\begin{IEEEbiography}[{\includegraphics
[width=1in,height=1.25in,clip,
keepaspectratio]{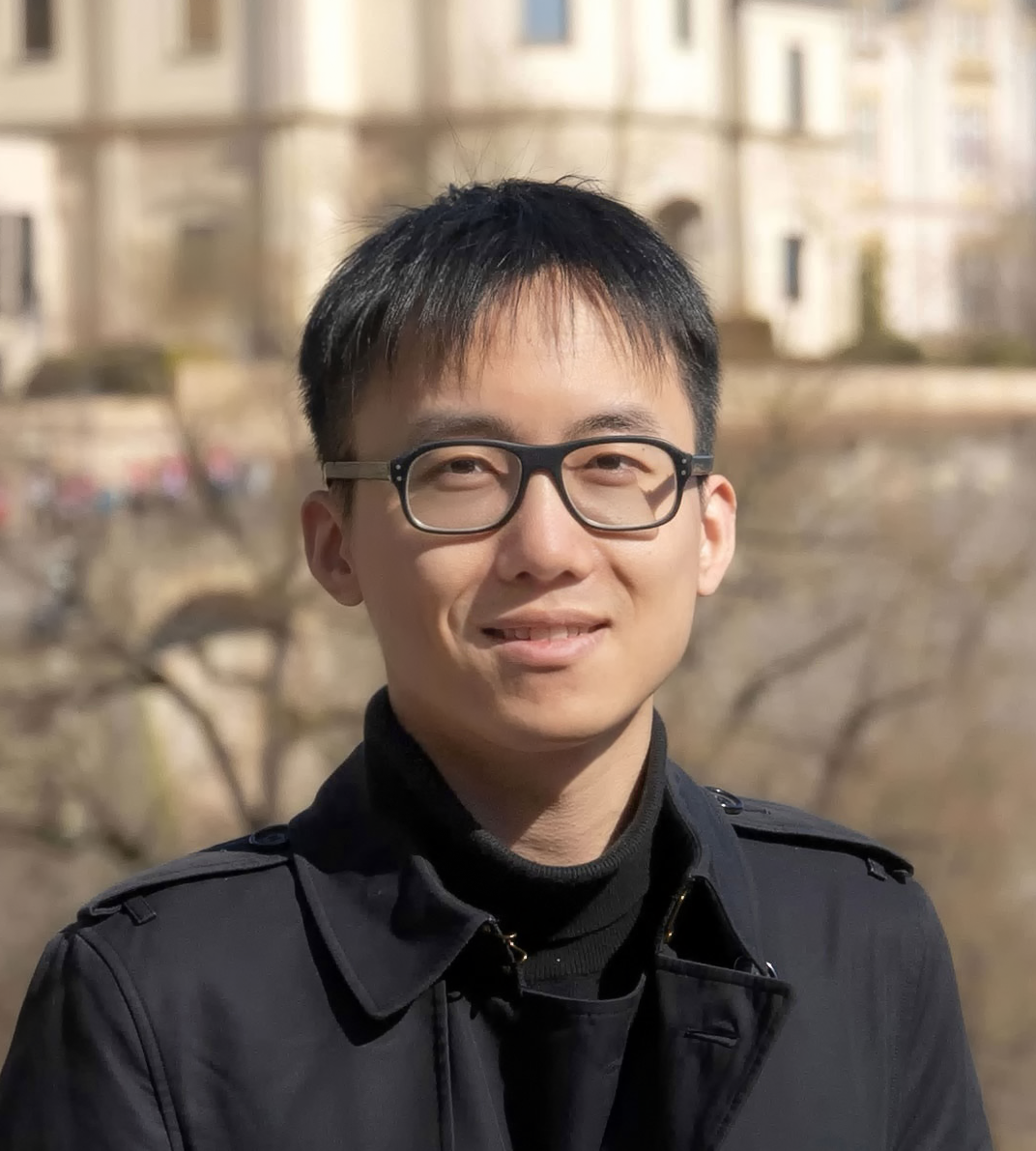}}]
{Xingang Pan}
Xingang Pan received the PhD degree in information engineering from The Chinese University of Hong Kong in 2021. He is currently a Nanyang assistant professor at College of Computing and Data Science, Nanyang Technological University. Previously, he was a postdoctoral researcher at Max Planck Institute for Informatics from 2021 to 2023. His research interests include generative models, visual generation, and 3D vision. He serves as an Area Chair for CVPR and 3DV. He is awarded the Singapore NRF Fellowship in 2024.
\end{IEEEbiography}

\begin{IEEEbiography}[{\includegraphics
[width=1in,height=1.25in,clip,
keepaspectratio]{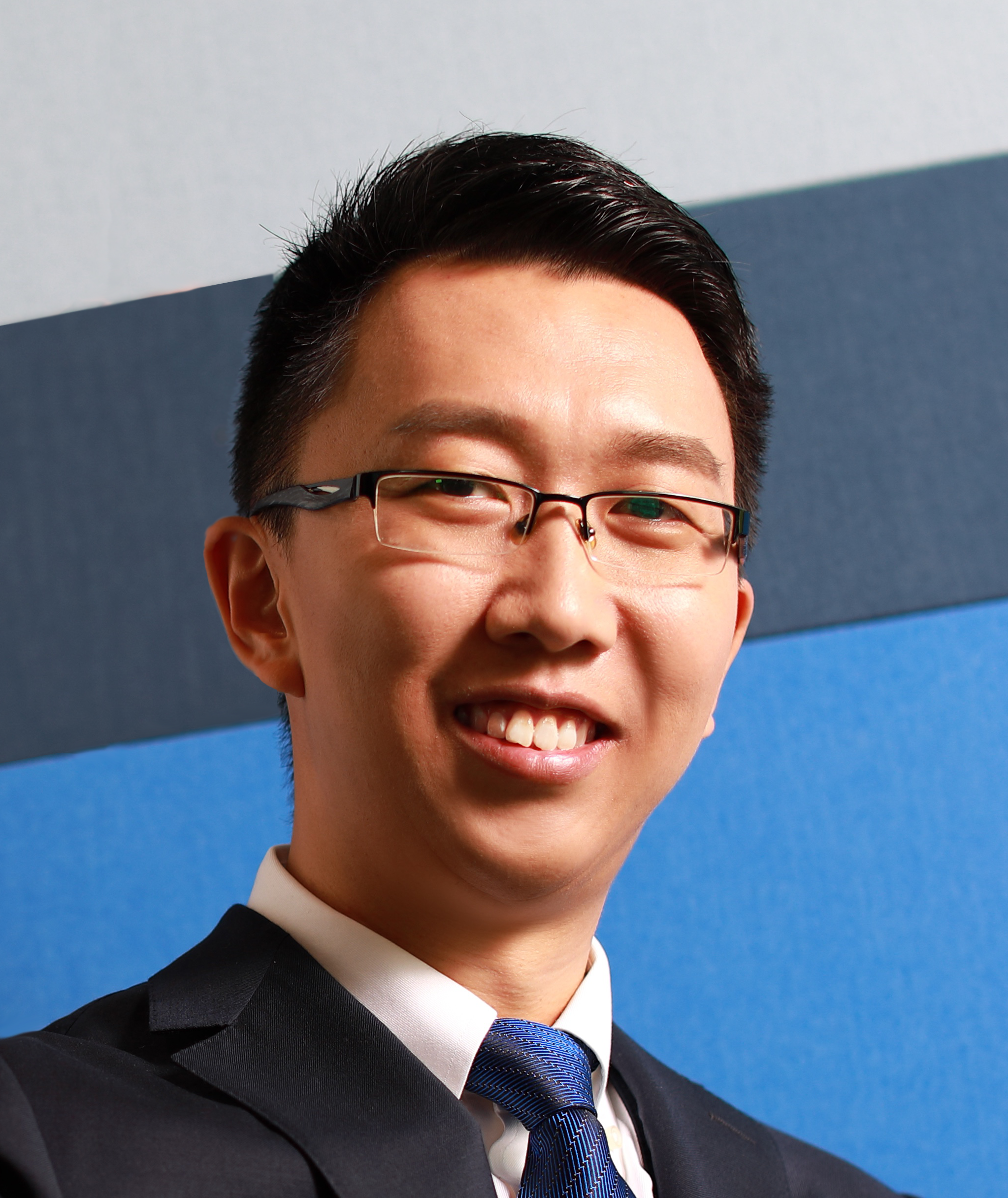}}]
{Chen Change Loy} (Senior Member, IEEE) is a President's Chair Professor at the School of Computer Science and Engineering, Nanyang Technological University, Singapore. Before joining NTU, he served as a Research Assistant Professor at the MMLab of The Chinese University of Hong Kong, from 2013 to 2018. He received his Ph.D. (2010) in Computer Science from the Queen Mary University of London. He was a postdoctoral researcher at Queen Mary University of London and Vision Semantics Limited, from 2010 to 2013.
He serves/served as an Associate Editor of the IEEE Transactions on Pattern Analysis and Machine Intelligence, International Journal of Computer Vision and Computer Vision and Image Understanding. He also serves/served as an Area Chair of major conferences such as ICCV, CVPR, ECCV, NeurIPS, and ICLR. He serves as the Program Co-Chair of CVPR 2026 and General Co-Chair of ACCV 2028.
His research interests include image/video restoration and enhancement, generative tasks, and representation learning.
\end{IEEEbiography}



\vfill

\clearpage

{\small
\bibliographystyle{ieee_fullname}
\bibliography{bibs/cvpr24.bib}
}

\end{document}